\documentclass[journal]{IEEEtran}
\ifCLASSOPTIONcompsoc
  \usepackage[nocompress]{cite}
\else
  \usepackage{cite}
\fi

\usepackage{microtype}
\usepackage{graphicx}

\usepackage[utf8]{inputenc} 
\usepackage[T1]{fontenc}    
\usepackage{url}            
\usepackage{amsfonts}       
\usepackage{nicefrac}       
\usepackage{microtype}      
\usepackage{graphicx}
\usepackage{amsmath}
\usepackage[export]{adjustbox}
\graphicspath{{TNNLS/images/}}
\usepackage{subcaption}
\usepackage{adjustbox}
\DeclareMathOperator*{\argmax}{arg\,max}

\usepackage[colorinlistoftodos,prependcaption,textsize=tiny]{todonotes}

\usepackage{algorithm2e}
\usepackage{algorithmic}
\usepackage{xcolor}
\begin{document}
%
\title{Deep Reinforcement Learning with Modulated Hebbian plus Q Network Architecture}
%
%
%
%

\author{Pawel~Ladosz,
Eseoghene~Ben-Iwhiwhu,
Jeffery~Dick,
Nicholas~Ketz,
Soheil~Kolouri,
Jeffrey~L.~Krichmar,
Praveen~Pilly,
Andrea~Soltoggio,

\thanks{This material is based upon work supported by the United States Air Force Research Laboratory (AFRL) and Defense Advanced Research Projects Agency (DARPA) under Contract No. FA8750-18-C-0103. Any opinions, findings and conclusions or recommendations expressed in this material are those of the author(s) and do not necessarily reflect the views of the United States Air Force Research Laboratory (AFRL) and Defense Advanced Research Projects Agency (DARPA). This work was also supported by Basic Science Research Program through the National Research Foundation of Korea (NRF) funded by the Ministry of Education (2020R1A6A1A03040570)}
\thanks{Pawel Ladosz was with Department of Computer Science, Loughborough University, Loughborough LE11 3TU, UK. He is now with School of Mechanical and Nuclear Engineering, UNIST, Ulsan 44919, South Korea E-mail: pladosz@unist.ac.kr}
\thanks{Eseoghene Ben-Iwhiwhu, Jeffery Dick and Andrea Soltoggio are with the Department of Computer Science, Loughborough University, Loughborough, UK.}
\thanks{Nicholas Ketz and Praveen Pilly are with with Information and Systems Sciences Laboratory, HRL Laboratories, 3011 Malibu Canyon Road, Malibu, CA 90265, USA}
\thanks{Soheil Kolouri is with the Computer Science Department at Vanderbilt University, Nashville, TN, 37235. This research was performed when he was with the Information and Systems Sciences Laboratory, HRL Laboratories, Malibu, CA, 90265}
\thanks{Jeffrey L. Krichmar is with Department of Cognitive Sciences, Department of Computer Science, University of California, Irvine, Irvine, CA, 92697, USA}
\thanks{Manuscript submitted to Special Issue on Biologically learned/inspired methods for sensing, control and decision making, October, 31, 2020}}

%
%

\markboth{Transactions on Neural Networks and Learning Systems}%
{Ladosz \MakeLowercase{\textit{et al.}}: Deep Reinforcement Learning with Modulated Hebbian plus Q Network Architecture}
%



\maketitle

\begin{abstract}
In this paper, we consider a subclass of partially observable Markov decision process (POMDP) problems which we termed confounding POMDPs. In these types of POMDPs temporal difference (TD)-based RL algorithms struggle, as TD error cannot be easily derived from observations. We solve these types of problems using a new bio-inspired neural architecture that combines a modulated Hebbian network (MOHN) with DQN, which we call modulated Hebbian plus Q network architecture (MOHQA). The key idea is to use a Hebbian network with rarely correlated bio-inspired neural traces to bridge temporal delays between actions and rewards when confounding observations and sparse rewards result in inaccurate TD errors. In MOHQA, DQN learns low-level features and control, while the MOHN contributes to the high-level decisions by associating rewards with past states and actions. Thus the proposed architecture combines two modules with significantly different learning algorithms, a Hebbian associative network and a classical DQN pipeline, exploiting the advantages of both. Simulations on a set of POMDPs and on the Malmo environment show that the proposed algorithm improved DQN's results and even outperformed control tests with advantage-actor critic (A2C), Quantile regression DQN with long short term memory (QRDQN+LSTM), Monte-Carlo policy gradient (REINFORCE) and aggregated memory for reinforcement learning (AMRL) algorithms on most difficult POMDPs with confounding stimuli and sparse rewards.
\end{abstract}


%
\IEEEpeerreviewmaketitle

\section{Introduction}
\IEEEPARstart{I}{n} this paper we address a sub-class of partially observable Markov decision processes (POMDPs) problems which we termed \emph{confounding} POMDPs. Confounding POMDPs are characterised by two aspects which are problematic for reinforcement learning (RL) algorithms: i) observations from the same set occur at random locations in the POMDP (we call such observations confounding observations) and ii) both positive and negative rewards are rare. Note that not all POMDPs are confounding, for example, using a single frame from Atari Breakout as a state is a POMDP as the agent has no information of the ball's direction and velocity. Non-confounding POMDPs can be solved with a memory system. In Atari Breakout one can stack multiple frames to derive an markov decision process (MDP) state\cite{Mnih2013PlayingAW}. However, in confounding POMDPs even memory-based approaches fail\cite{Wierstra2007} because the histories of observations do not repeat (due to observations occurring at random locations through POMDP), thus the underlying state cannot be easily inferred from observations. The sparsity of the reward implies that many such uninformative observations occur in between rewards. The result is that TD errors cannot be computed if the state cannot be derived either from the present observation or a history, or alternatively, they become quickly inaccurate due to sparse rewards. Confounding observations and sparse rewards are common in many scenarios for example, when driving across a town from point A to B, cars parked to the side of the road are not useful landmarks, whereas junctions or buildings are. Cars parked on the side of the road are creating confounding observations as they occur randomly in the problem and are irrelevant to a task of getting from A to B. Also, in this example, the reward might be rarely provided, i.e. not at each correct turn, but only when the destination is reached.

One way to apply RL to POMDP is to try to derive a state from the history of observations, and thus LSTM has been employed in ~\cite{hausknecht2015deep, heess2015memory, mirowski2016learning, Wierstra09}. Another notable LSTM-based approach was shown in~\cite{Zhu2017OnID}, where action and states are combined and inserted into the network as a single input resulting in more observable states. LSTMs were even used with hierarchical reinforcement learning to solve some types of POMDPs~\cite{Le2018}. One of the more recent approaches called AMRL\cite{Beck2020} combines LSTM-based memory with a latent space averaging over time to boost the signal to noise ratio in the latent space to solve POMDP problems. To deal with sparse rewards (in a non-POMDP setting)~\cite{Stepleton2018LowpassRN}, proposed the use of “recurrent neural networks (RNNs) with concrete”, which remembers states for longer by using multiple RNNs connected in series. ~\cite{Parisotto17} used differential neural computing to solve POMDPs requiring memory. Other memory types, which do not rely on neural networks, were also developed. For example, in~\cite{Oh2016} the agent stores a certain number of past states and retrieves relevant states when necessary. Zhang et al.~\cite{Zhang2016} introduced continuous memory states to better deal with problems in the continuous domain. Such memory-based approaches, while effective in many cases, suffer in cases in which observations derive from a large set, and thus histories do not repeat. While the training phase for an LSTM could provide the network with the ability to discard confounding observations, in practice this is often hard to achieve. 

There exist other methods that do not rely on memory, for example, ~\cite{Steckelmacher2018ReinforcementLI} introduced option-observation initiation sets, to work with options from~\cite{Precup00temporalabstraction} to make them suitable for solving POMDPs. Another method is based on the well known probabilistic inference for learning control (PILCO) algorithm\cite{Deisenroth2011} in~\cite{McAllister2017}. That work proposes two additions for PILCO:  i) direct training and ii) filtering to deal with noisy state space. Generative models were used to update beliefs about the environment in ~\cite{Igl2018}. These approaches do not suffer from non-repeating histories, but require dense rewards.

To address the challenges of solving confounding POMDPs, we propose an approach called modulated Hebbian plus Q network architecture (MOHQA). The architecture is composed of two parts: a standard DQN network, and a layer of reward-modulated Hebbian network\cite{hebb2005organization,soltoggio2013solving} with neural eligibility traces (MOHN), parallel to the Q-network head that contributes to decision making. The combination of two learning modules in one architecture differs from  Backpropamine ~\cite{miconi18a, Miconi2020}. In Backpropamine, Hebbian learning and backpropagation are used on the same connection, where backpropagation determines the utilization of the Hebbian component. In Backpropamine, Hebbian component does not result in a permanent weight increase and is nullified at the end of each episode. In essence Hebbian connections are used for a within-episode memory, while backpropagation is used to learn scaling factors, thereby deciding when and how much of a Hebbian component to use within episodes. On the other hand, in MOHQA, Hebbian components result in a permanent weight increase. Also MOHQA is utilizing sparse correlations. Moreover, Backpropamine has only been shown to work in a very limited setting in reinforcement learning with a single layer network and a 1-dimensional state space.

In the proposed architecture, DQN acts as a feature provider via convolutional layers to the MOHN, and contributes to the final decision in equal measure as the parallel MOHN.
The addition of MOHN helps DQN discriminate between pivotal decision points and confounding observations, thus gaining a learning advantage in confounding POMDP problems. MOHN uses two key mechanisms: i) rarely correlated eligibility traces to associate a state-action pair with reward and ii) Hebbian learning for rapid learning from few examples. The ability of the modulated Hebbian-like network with rare eligibility traces to bridge temporal gaps between events and rewards and rapid learning from few examples was demonstrated in a spiking neural model in Izhikevich \cite{izhikevich2007solving}, and an equivalent model for rate-based neurons was shown effective in simulations and robotics applications \cite{soltoggio2013solving,soltoggio2013rare,soltoggio2013learning}. 

Eligibility traces, which is one of MOHQA's key mechanisms, is  not new in reinforcement learning \cite{sutton2018reinforcement,Munos2016}. However, eligibility traces in MOHQA use forward view and implement a rare (or sparse) correlation mechanism that distinguishes MOHQA's traces from traditional reinforcement learning ones, such as SARSA($\lambda$), Q($\lambda$) or Monte Carlo methods (which are special case of eligibility traces approach when $\lambda=1$) such as REINFORCE. While forward view and backward view eligibility traces are largely equivalent, rare correlations give MOHQA two distinct advantages. Rare correlation means only a small percentage of all weights are considered for an update for a given state-action pair. This means that, compared to traditional reinforcement learning, MOHQA (i) has the increased ability to perform well with pure rewards signals, (ii) has the ability to cope with noise in the feature space (a problem that has been recently discussed in great detail in~\cite{Beck2020}), by ignoring the noisy inputs and only focusing on significant input features.



\begin{figure*}
   \centering
    \begin{subfigure}[t]{0.45\textwidth}
        \centering
        \includegraphics[width=\textwidth]{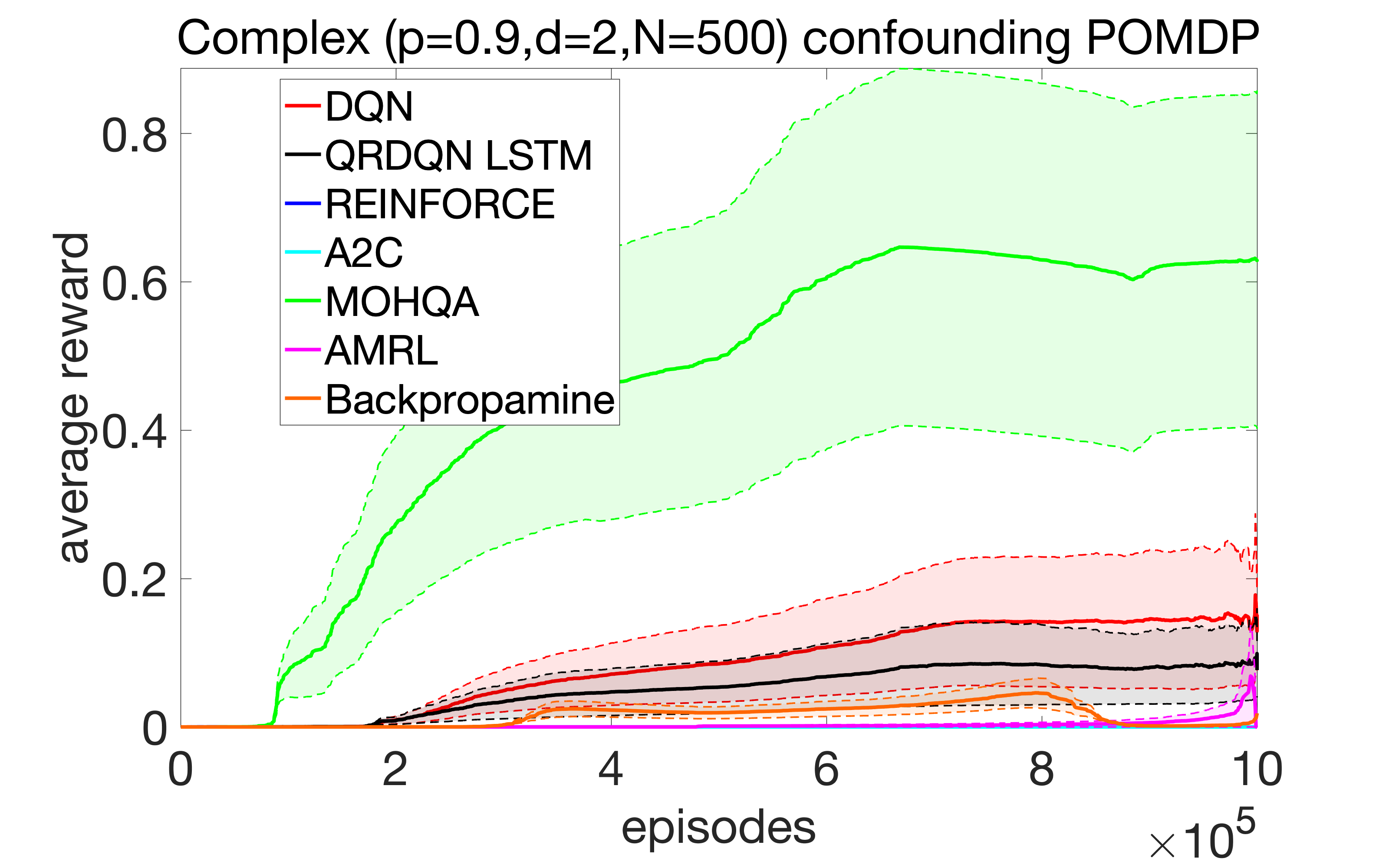}
        \caption{} 
        \label{fig:complex_3_intro}
    \end{subfigure}
    \begin{subfigure}[t]{0.45\textwidth}
        \centering
        \includegraphics[width=\textwidth]{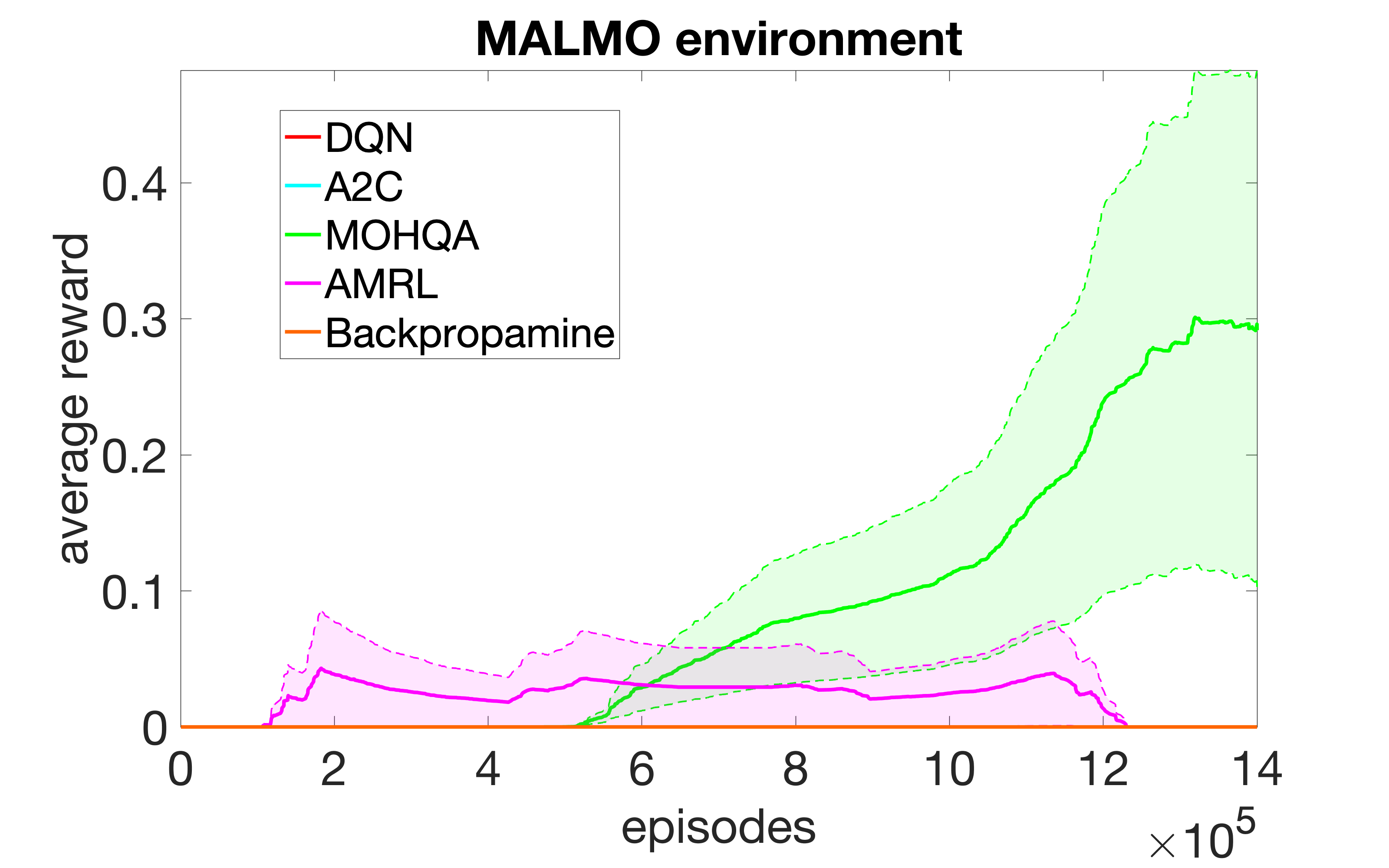}
        \caption{}
        \label{fig:complex_Malmo}
    \end{subfigure}
\caption{Selected performance of MOHQA on benchmarks considered in this paper. (a) Comparison of results in a very complex POMDP set in CT-graph. MOHQA outperforms all other baselines by at least 33\%, (b) Comparison of results in Malmo benchmark. MOHQA is outperforming AMRL, DQN and A2C by at least 33\%.}
\label{fig:results_intro}
\end{figure*}

The architecture is tested on a set of generalized reward-based decision problems that include confounding POMDPs. Tests include comparisons with a baseline DQN \cite{Mnih2013PlayingAW}, QRDQN+LSTM \cite{hausknecht2015deep}, REINFORCE, A2C \cite{mnih2016asynchronous} and AMRL \cite{Beck2020}. Our simulations show that MOHQA can solve confounding POMDPs which DQN cannot, and is even able to outperform A2C, memory-based AMRL and REINFORCE by at least 33\% in more complex scenarios (see Fig.~\ref{fig:results_intro}). To the best of the authors' knowledge, the proposed approach is the first application of a combined modulated Hebbian and DQN network to improve learning in the presence of partially observable Markov states.

\section{Confounding POMDPs problems}

In this work, confounding POMDPs are represented as environments where leading-to-reward decision points, occur occasionally and are separated by confounding wait states where a fixed policy is required, e.g., wait action. This is a fairly common case in robotics and games. In the driving analogy we made above, key decision points are at junctions, while wait states are those along straight segments where parked cars are confounding stimuli. The challenge of the confounding POMDPs derives from i) a large number of confounding observations and ii) the sparsity of the reward. Here we encode those type of problems within two benchmarks i) a graph-based benchmark with 2d synthetic inputs we named configurable tree graph (CT-graph) and ii) a well known 3D benchmark, Malmo \cite{Oh2016, Beck2020}.

\subsection{CT-graph Environment}

\begin{figure*}
   \centering
    \includegraphics[width=0.95\textwidth]{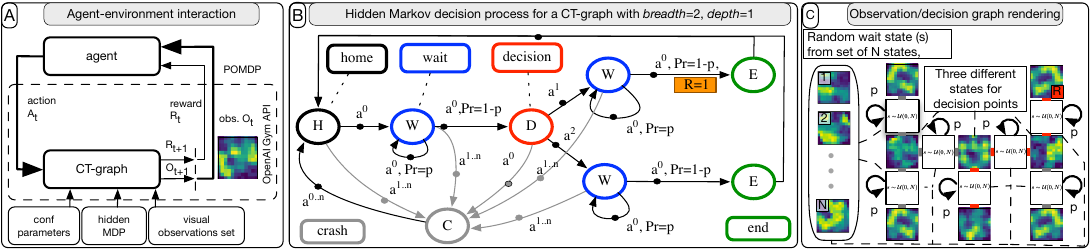}
   \caption{The CT-graph problem. (A) The agent-environment interaction is similar to the standard RL with the difference that the environment provides observations. (B) Representation of the hidden Markov process: the wait-action $a^0$ is required at wait states, while act-actions $a^1$ and $a^2$ are required at decision points. The panel shows a unit (with one branching point) that can be repeated to obtain arbitrarily large problems. (C) Graphical representation of the observations and decision for a CT-graph of depth 2, i.e., two sequential branching points.}
\label{fig:problem}
\end{figure*}

The CT-graph is an abstraction of the decision-making process and it allows designing confounding POMDPs with the measurable complexity and partial observability in quantifiable metrics (see Fig.~\ref{fig:problem} for pictographical representation). A CT-graph can be thought of as a tree-like decision graph with two types of nodes (i) \emph{decision states} (where the CT-graph branches into multiple sub-trees) and (ii) \emph{wait states}  (where the tree does not branch). Actions are also of two types: \emph{wait-actions} and \emph{act-actions}. 

A reward is located at one particular leaf node in the tree graph (Fig.~\ref{fig:problem}(C)). The agent is required to perform wait-actions while in a wait state. While while at a decision point the agent chooses from a set of act-actions, where the specific act-action chosen determines the path that the agent follows along the tree. Wait states lead to themselves with a delay probability $p$ or to the next state in the sequence with probability $1-p$. At each decision point there are $b$ options corresponding to the branches in the sub-trees of a particular decision point. The number of decision points between home and reward state is the depth $d$ of the tree graph. The environment has an optimal sequence of actions that leads to one unique leaf of the tree graph that returns a reward of one. The branch factor $b$, the depth $d$, the delay probability $p$, and the sets of observations are configurable parameters, making this problem a blueprint for a large set of benchmarks, from simple to extremely difficult problems for medium to large size graphs (Fig. \ref{fig:problem}).  

\paragraph{CT-graph as a confounding POMDP problem.} The specific configurations of the CT-graph that realise confounding POMDPs are where wait state observations are chosen randomly from a large set $N$ and reward is sparse (the delay probability $p$ is high). Such a setup represents a decision-making process in which confounding observations (wait states) are more common than decision points. With this property, we observed that RL agents cannot easily learn optimal policies as histories do not repeat and TD error estimates are inaccurate.

\subsection{Malmo Environment}

\begin{figure}
   \centering
    \includegraphics[width=0.25\textwidth]{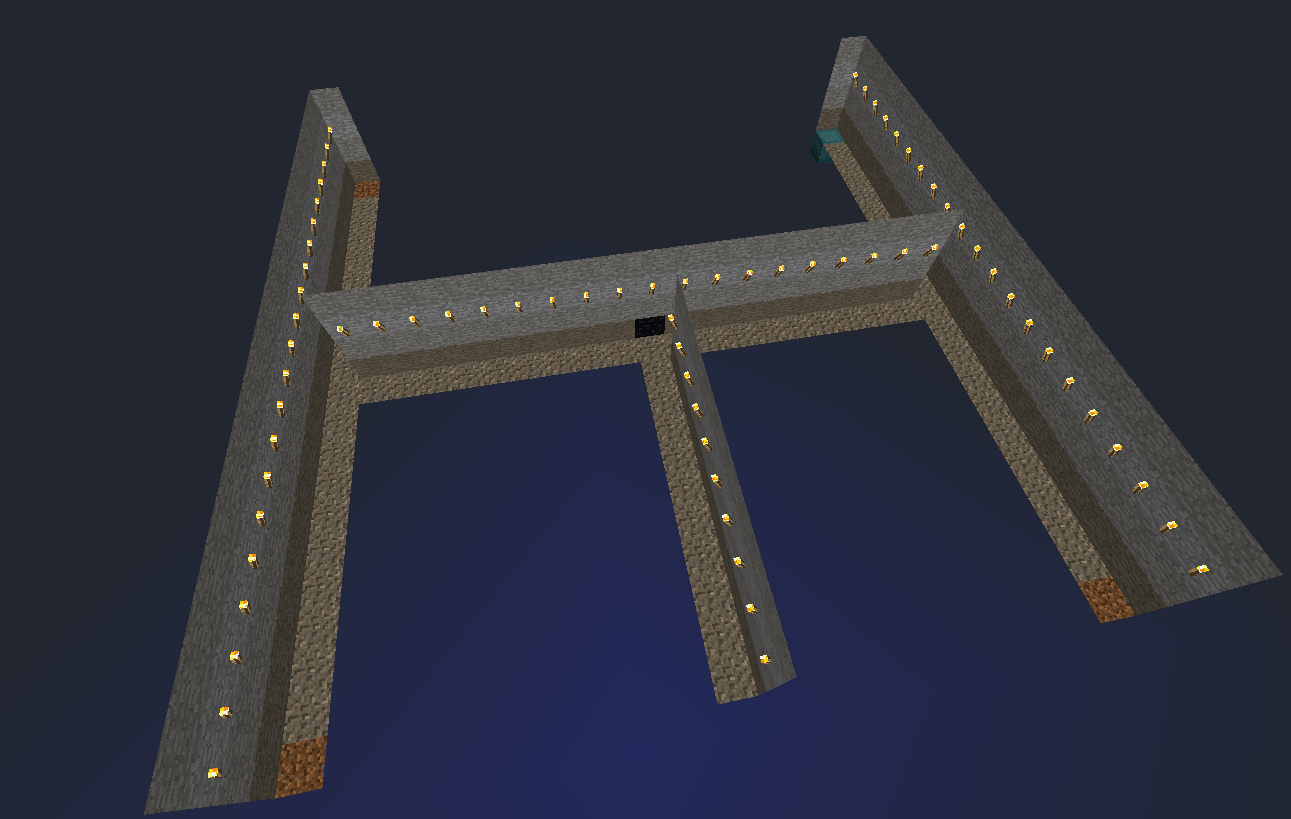}
   \caption{Bird's eye view of maze used in this paper. The structure of the maze with two decision points and four endings can be noticed. The teal block signifies position of the reward.}
\label{fig:minecraft_overview}
\end{figure}

Malmo~\cite{Johnson2016} is an environment based on the 3D game Minecraft. The environment (Fig.~\ref{fig:minecraft_overview}) is configured with a similar outline as the CT-graph, with junctions (similar to decision states) which are separated by long corridors (similar to wait for states). The reward is positioned at one leaf node (see a teal square in Fig.~\ref{fig:minecraft_overview}). The agent receives no reward or punishment for every other step. 

At each step, the agent makes a discrete move either left, right or forward. The forward action is equivalent to the wait-action of the CT-graph while left/right are decisions. To help with exploration speed, the backward action is disabled and a step forward after each turn action is performed. Stepping forward after each turn means: i) the agent can only take one turn in a decision point; ii) if a left/right action is taken in a corridor, the episode is terminated. Those two simplifications stopped the agent from rotating on the spot during exploration.

\paragraph{Malmo as a confounding POMDP problem.} The problem in Malmo is a confounding POMDP as the some corridor states look similar to each other regardless of the location in the maze, and therefore affect the ability to compute the TD-error. For example, the corridor after the first turning point looks the same regardless of whether the agent chose the left or right action. While histories are easier to use here than in the CT-graph, it is still challenging to extract what is useful to remember due to the same observations appearing at different places in the problem space. 

For more details about the environments please see Appendix 3.

\section{The modulated Hebbian plus Q network architecture (MOHQA)}

In this section a brief description of MOHQA architecture is provided, followed by a description of two key components in MOHQA: DQN and modulated Hebbian network (MOHN) and their integration.

\begin{figure*}[h]
   \centering
    \includegraphics[width=0.74\textwidth]{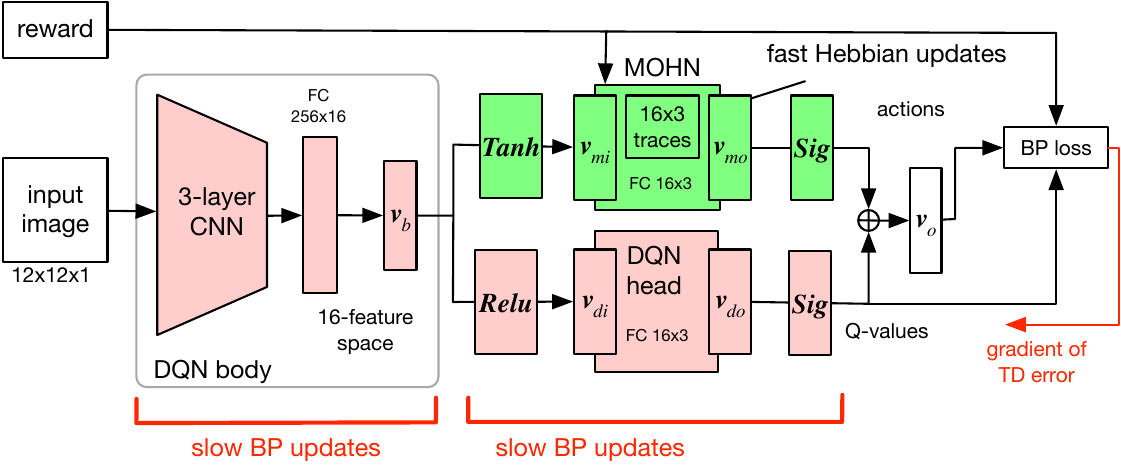}
   \caption{Graphical representation of the MOHQA. The DQN body feeds the feature space to both the DQN head and the MOHN. The action-value vector is the sum of the outputs of the two heads. The learning in the MOHN is regulated by modulated Hebbian plasticity. The learning in the DQN network is performed by back propagation.}
\label{fig:MOHQA}
\end{figure*}

MOHQA is composed of two main parts: a deep Q-network \cite{mnih2015human} (DQN), and a modulated Hebbian network (MOHN) that is plugged into the Q-network as a parallel unit to the DQN head (Fig. \ref{fig:MOHQA}). DQN's main contribution is providing high-level features to MOHN, while MOHN's main contribution is decision making.

\subsection{DQN description}
MOHN is a one layer network which is intended to work on high level features, thus it needs a 'supplier' of high level features. In this work DQN is used but, in theory, any deep reinforcement learning approach could have been used instead of DQN.
The DQN follows the implementation from \cite{mnih2015human} \footnote{We used the implementation of that work from Shangtong Zhang URL: https://github.com/ShangtongZhang/DeepRL}.
A DQN is used to approximate the optimal Q-value function, defined as 
\begin{equation}
\begin{aligned}
Q(s,a)^\star= & \max_{\pi}\mathbb{E} [r_t+\gamma r_{t+1}+\gamma^2 r_{t+2} \\
              &+... | s_t=s, a_t=a,\pi], \\
\end{aligned}
\end{equation}
where $r_t$ is the reward at time $t$, $s_t$ is the state at time $t$, $a_t$ is the action at time $t$ and $\pi$ is the policy. 
To solve instability issues caused by representing action-value pairs with the network, Minh et al.,~\cite{mnih2015human} proposed two innovations i) experience replay and ii) a target network that is only periodically updated. Note, $\Phi^-$ and $\Phi$ denote set of parameters of target and predictor networks respectively. The parameters $\Phi^-$ are updated with $\Phi$ every $K$ steps. The output of the DQN body is a set of features that are used as input to both the MOHN and DQN heads. The layers and sizes of the DQN body are summarized in supplementary Material 1.

Both the DQN head and the DQN body are trained by back propagation to minimize the TD error \cite{mnih2015human}.

\subsection{MOHN Description}

In this section discussion of key mechanisms allowing MOHN to solve confounding POMDPs and its formulation is presented. 

\subsubsection{Associating stimulus-action pairs with distal rewards by means of MOHN}
The MOHN in this work is an adaptation of a bio-inspired, unsupervised and modulated Hebbian network proposed in \cite{izhikevich2007solving,soltoggio2013solving}.
In those studies, a Hebbian modulated network was shown to cope with two challenges of confounding POMDPs --- sparse rewards and confounding stimuli --- outside a RL framework, i.e. when neither states nor TD errors are defined. It is worth reiterating that, in confounding POMDPs, the TD error computation is inaccurate and cannot be used to propagate the Q-values. Moreover, even memory does not help to solve the problem as the history of observations maps to a large space due to stochastic observations, making learning history of action-observations ineffective. The MOHN solves sparse reward and confounding stimuli problems by using rarely correlated eligibility traces and Hebbian learning. Eligibility traces do not rely on TD-error computation, instead they directly associate activation of certain neurons with obtaining rewards later; while rare correlation allows MOHN to cope with noise in inputs, grow weights without instability and utilize all the weights efficiently. Additionally, as MOHN is utilizing Hebbian learning, it can have a high learning rate without instability issues. The result is that observations-action pairs are effectively associated with later rewards by means of traces, weights responsible for an action are updated rapidly, and intervening events between actions and rewards are ``ignored''. 

\subsubsection{MOHN learning mechanism using STDP-inspired plasticity and neural eligibility traces}
The learning mechanism in MOHN can be briefly summarized as follows. An action taken in a state triggers a number of traces for weights between highly correlated input/output neurons. The traces are decayed exponentially with each step. At each step, weights are either decreased if the agent did not score or increased if the agent scored. This mechanism enables the network to find the correct association between actions and rewards with intervening confounding stimuli, thus avoiding shortcomings of propagating TD-error.

To be more specific, neural eligibility traces generation mechanisms use two principles: (i) causal relationships between observation and actions, and (ii) sparse correlations. The causal relationships are derived by applying the Hebbian multiplication rule to successive, rather than simultaneous, simulation steps, so that Hebbian terms capture the contribution of presynaptic activity to the activity of a postsynaptic neuron, similarly to the Spike-timing-dependent plasticity (STDP) rule. This is also sometimes called an asymmetrical learning window \cite{kempter1999hebbian}. Sparse correlations are explicitly imposed by selecting the top $\theta\%$ and bottom $\theta\%$ of correlations/decorrelations that loosely map the concept of the STDP time window to a rate-based model \cite{soltoggio2013rare}. A modified Hebbian term $\Theta$ between a presynaptic neuron $i$ and a postsynaptic neuron $j$ is updated according to the equation:
\begin{flalign}
\begin{split}
\Theta_{pre\to post}(t) = \qquad  \qquad \qquad \qquad \qquad \qquad \qquad \ \ \ \ \\
\left\{
\begin{array}{ll}
1 \, & \mathrm{if}\,  v_{i} \cdot v_{j}\ \mathrm{is\ in\ top\  \theta\%}, \\
-1 \, & \mathrm{if}\,  v_{i} \cdot v_{j}\ \mathrm{is\ in\ bottom\  \theta\%}, \\
0 \, & \textrm{otherwise},\\
\end{array} \right.
\\\label{eq.RCHP}
\end{split}
\end{flalign}
where $v_{i}$ is the output values of the presynaptic neurons equivalent to the input to the MOHN, $\mathbf{v}_{mi}$ (Fig. \ref{fig:MOHQA}), minus its own running average to enhance the detection of changes in the feature space and $v_{j}$ is defined as:
\begin{equation}
\mathbf{v}_{j} = \Gamma \big(\mathbf{v}_{mo} \big),
\label{eq.neuron}
\end{equation}
where $\mathbf{v}_{mo}$ is an output of MOHN head and $\Gamma$ is a function that returns a one-hot vector with the 1 value at the index of the maximum value (note that $v_{i}$ is used both in forward pass and weight updates, while $v_{j}$ is used only for weight update and $v_{mo}$ is used in forward pass). The one-hot function has the purpose of increasing the traces for the weights that are afferent to the action-triggering neuron. Finally, the neural eligibility traces matrix $E(t)$ is formulated based on $\Theta$ and a time decay constant $\tau_{E}$ as
\begin{eqnarray}
\dot{E}(t) &=& -E(t)/\tau_{E} + \Theta(t) \quad.
\label{eq.eTupdate}
\end{eqnarray} 
To update weights, the traces matrix $E(t)$ is multiplied with modulatory signal (reward plus a small baseline modulation, i.e., r(t) + $b$):
\begin{eqnarray}
\Delta\mathbf{w}(t) &=& (r(t) + b) \cdot E(t)\label{eq.wUpdate}.\label{eq.m}
\end{eqnarray}
The weights are clipped in the interval [-1,1] to contain Hebbian updates \cite{miller1994role,soltoggio2012modulated}. 



\subsection{Integration of DQN with MOHN}

To integrate DQN with MOHN it is necessary to devise an action selection mechanism between two head modules of Fig.~\ref{fig:MOHQA}. Thus, the action is chosen by combining the outputs of the two heads.

Outputs of both heads are combined to create the MOHQA Q-values, $\mathbf{v}_o$, which are defined as:
\begin{equation}
\mathbf{v}_{o} = \mathbf{v}_{mo} + \mathbf{v}_{do} = \mathbf{v}_{mo} + Q(s,A;\Phi^-_i),
\end{equation}
where $s$ indicates that an observation is used to approximate the state, even when this is incorrect due to partial observability.
Then action is chosen from MOHQA Q-values as follows:
\begin{equation}
a_b=\argmax_{a}(\mathbf{v}_o),
\end{equation}
To help DQN learn features in the desired state, the loss function uses the difference between best action as indicated by the Q-value of both DQN and MOHN and Q-value indicated by the DQN:
\begin{equation}
\begin{aligned}
L(\Phi)= & \mathbb{E}_{(s,a,r,s')~U(D)}\Big(r+\gamma \max_{a_b} \mathbf{v}_{o}(s',a_b, \Phi^-_i) \\
         & -Q(s,a;\Phi_i)\Big)^2, \\
\end{aligned}
\label{loss}
\end{equation}

where $\Phi^-$ are parameters from the target network and $\Phi$ are parameters from the prediction network. A summary of the full MOHQA algorithm is shown in Algorithm~\ref{MOHQA_algo}.

\begin{algorithm}
\SetAlgoLined
\KwIn{1 observation $x_i$}
 Initialize replay memory $D$ to capacity $N$; \\
 Initialize DQN with random weights and MOHN with zero weights \\
 \For{episode = $1,M$} {
  Initialise state $s_1={x_1}$ and preprocessed (normalized state) $\phi_t=\phi(s_1)$ \\
    \For{$t=1,T$} {
    Compute ${v}_o$ using $\phi(s_t)$ \\ 
    With probability $\epsilon$ select random action $a_t$ \\
    Otherwise select $a_t=\argmax_{a}(\mathbf{v}_o)$ \\
    Execute action $a_t$, observe reward $r_t$ and image $x_{t+1}$ \\
    Update eligibility traces using Eq.~\ref{eq.RCHP} and~\ref{eq.eTupdate} \\
    Update MOHN weights (Eq.~\ref{eq.m}) \\
    Set $s_{t+1}=x_{t+1}$  and preprocess $\phi_{t+1}=\phi(s_{t+1})$ \\
    Store transition ($\phi_t$, $a_t$, $r_t$, $\phi_{t+1}$) in $D$ \\
    \ForEach {$k$ steps} {
        Sample random minibatch of transitions $(\phi_j,a_j,r_j,\phi_j+1)$ from $D$\\
        Perform a gradient descent step according to Eq.~\ref{loss} \\}
    }
    Reset eligibility traces,
    reset averaging
}
 \caption{MOHQA}
 \label{MOHQA_algo}
\end{algorithm}
Where $x_t$ is observation from the environment and $\phi(\cdot)$ is a function normalizing color between 0 and 1.

\section{Results}
This section reports the analysis of how (i) learning mechanisms in MOHN compare to those of REINFORCE and DQN; (ii) the MOHN and new loss function enhances the features from DQN to solve the POMDP problems; (iii) the MOHQA compares against DQN, QRDQN+LSTM, REINFORCE, A2C and AMRL in the CT-graph and Malmo benchmarks.

\subsection{Comparison of MOHN learning to DQN and REINFORCE}

In this section, the MOHN's learning mechanisms i) Hebbian learning, ii) eligibility traces and iii) rare correlations are contrasted against two other classical learning methods i) TD learning in the form of DQN and ii) policy gradient in the form of REINFORCE. For this comparison, we train a one-layer network with each of the techniques and we remove the feature extraction part. Then, we feed a high-level one-hot feature space to one layer networks for the sole purpose of analysing the learning of the head modules. To be exact, observations are represented by 1-D vector of size $k$ on two different scenarios: i) with one element equal to $1$ and all others equal to $0$ to show the importance of eligibility traces and high learning rates and ii) with one element equal to 1 and all others assigned a random value using 10\% uniformly distributed random noise to show the importance of rare correlations. The singular element represents the unique observation (see Appendix 3 for state-space visualization). Moreover, the training is performed in an  imitation learning-like fashion, where an agent is led to the goal directly in each episode by an all-knowing oracle. Such a simple state space and imitation learning mean that the task is simplified to carry out the proposed analysis. All configuration parameters are provided in the supplementary material 1.

\begin{figure*}[htb]
\begin{tabular}{l|l|l}
Episode 2 & Episode 5 & Episode 10   \\ \hline
    \includegraphics[width=0.3\textwidth]{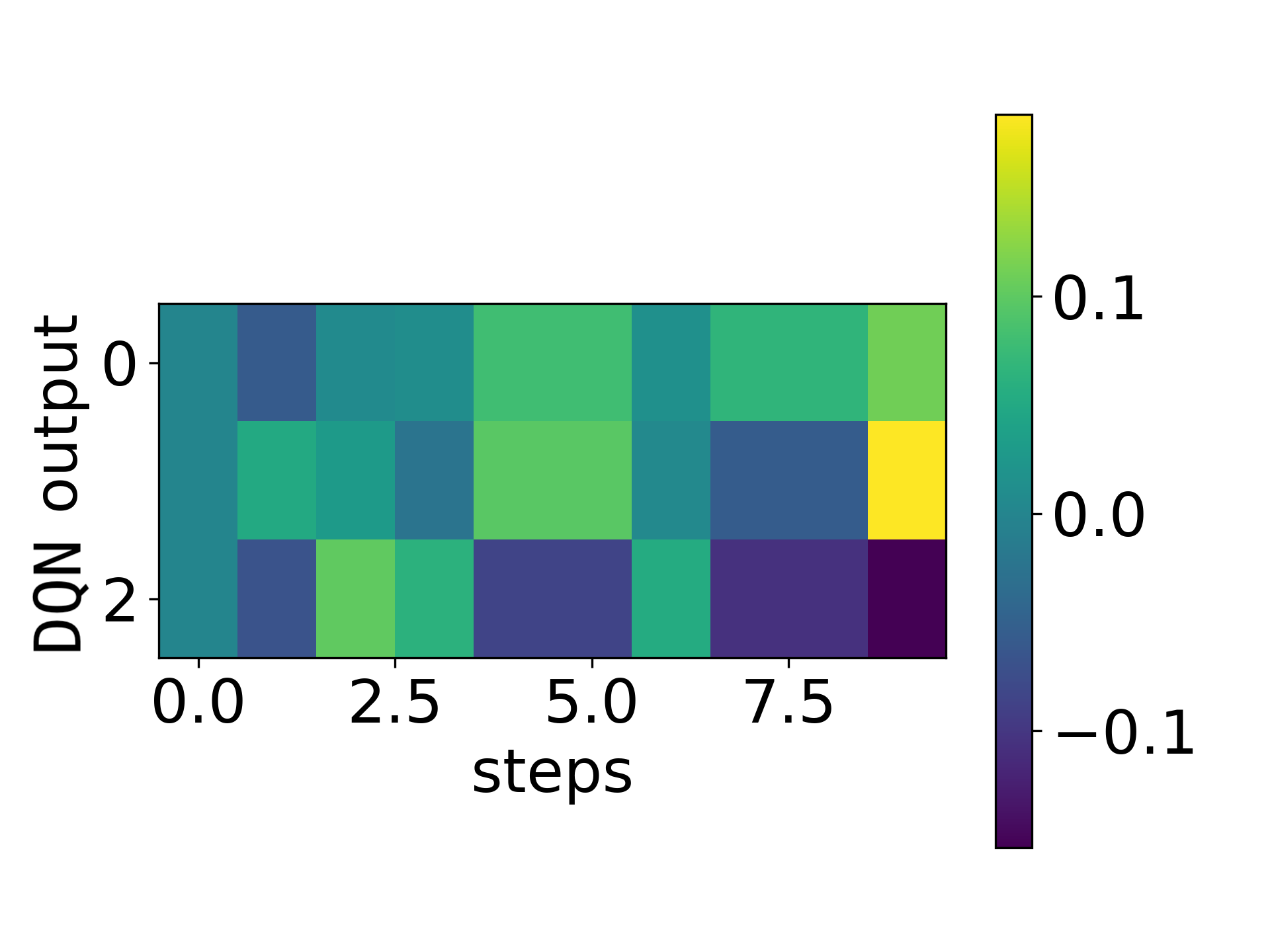}
 & 
    \includegraphics[width=0.3\textwidth]{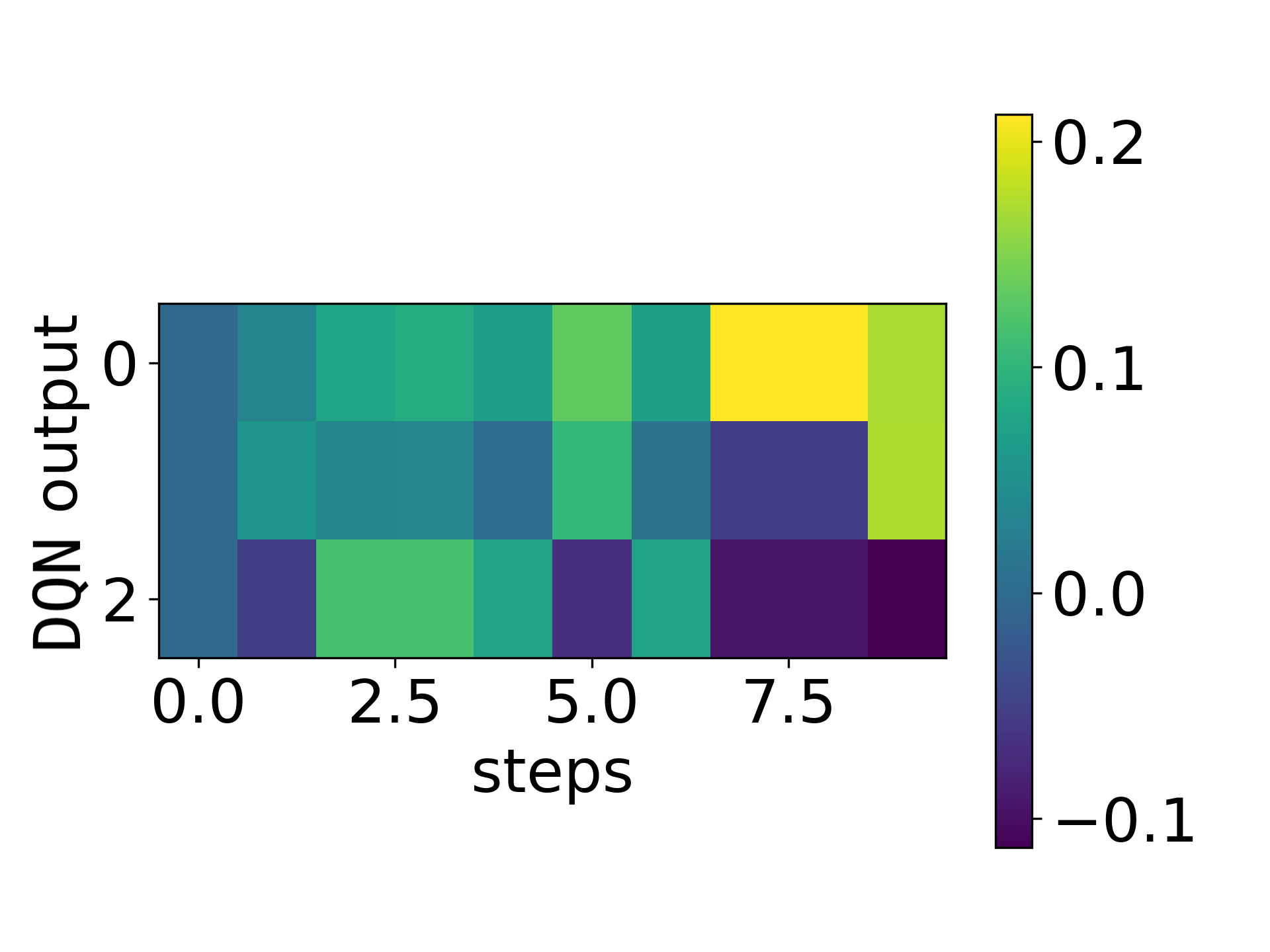}
& 
    \includegraphics[width=0.3\textwidth]{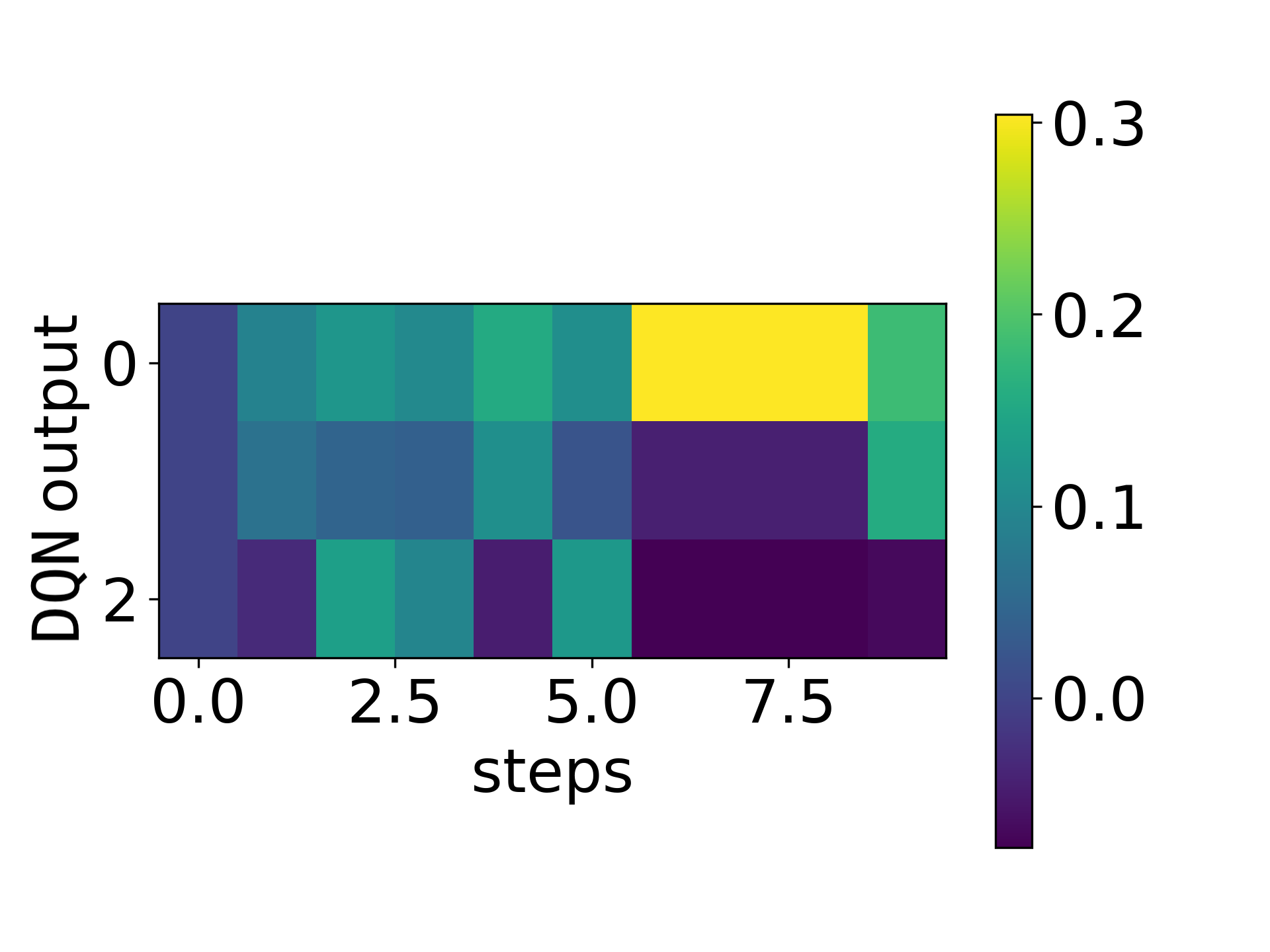}
 \\ 
    \includegraphics[width=0.3\textwidth]{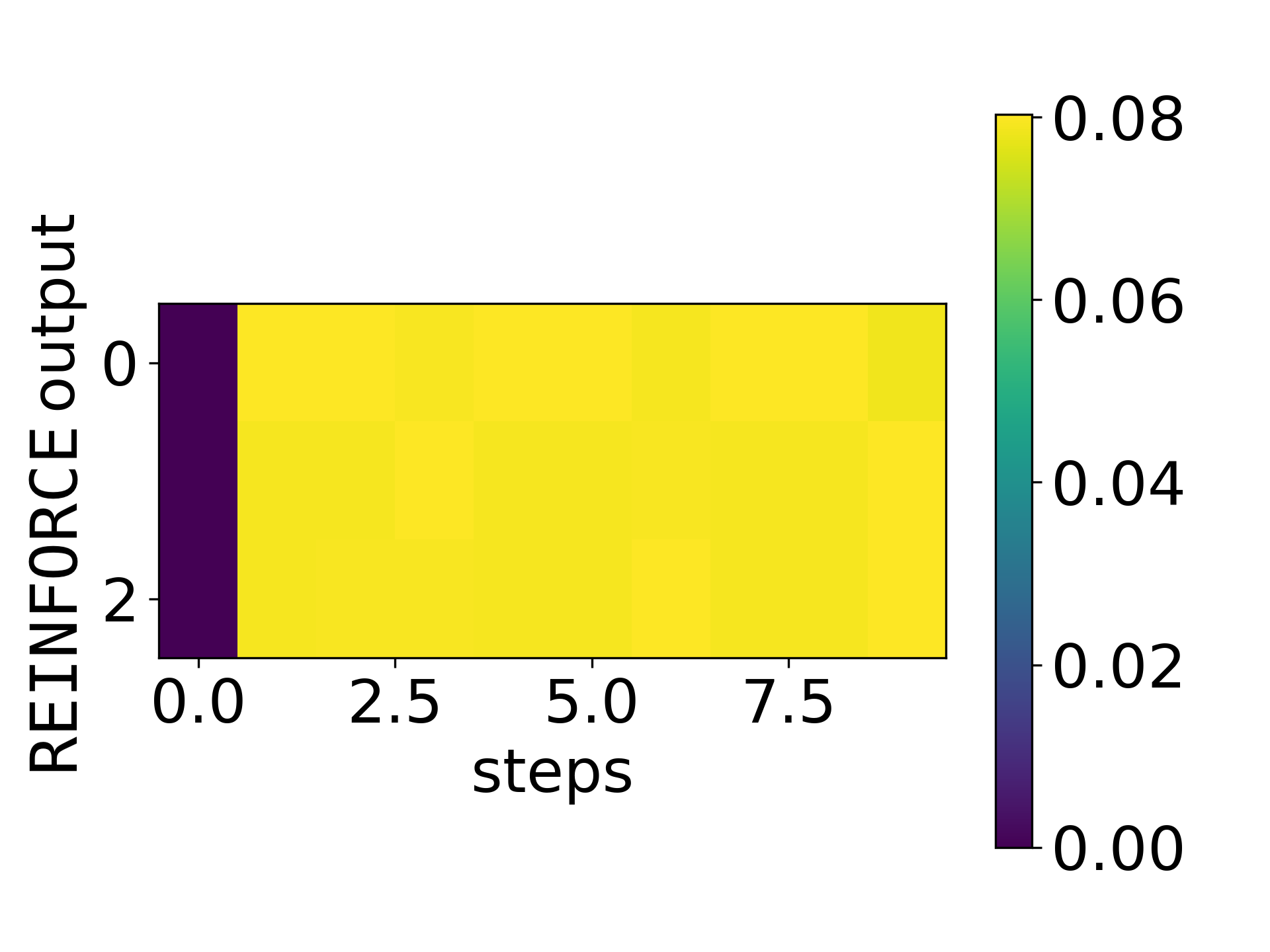}
 & 
    \includegraphics[width=0.3\textwidth]{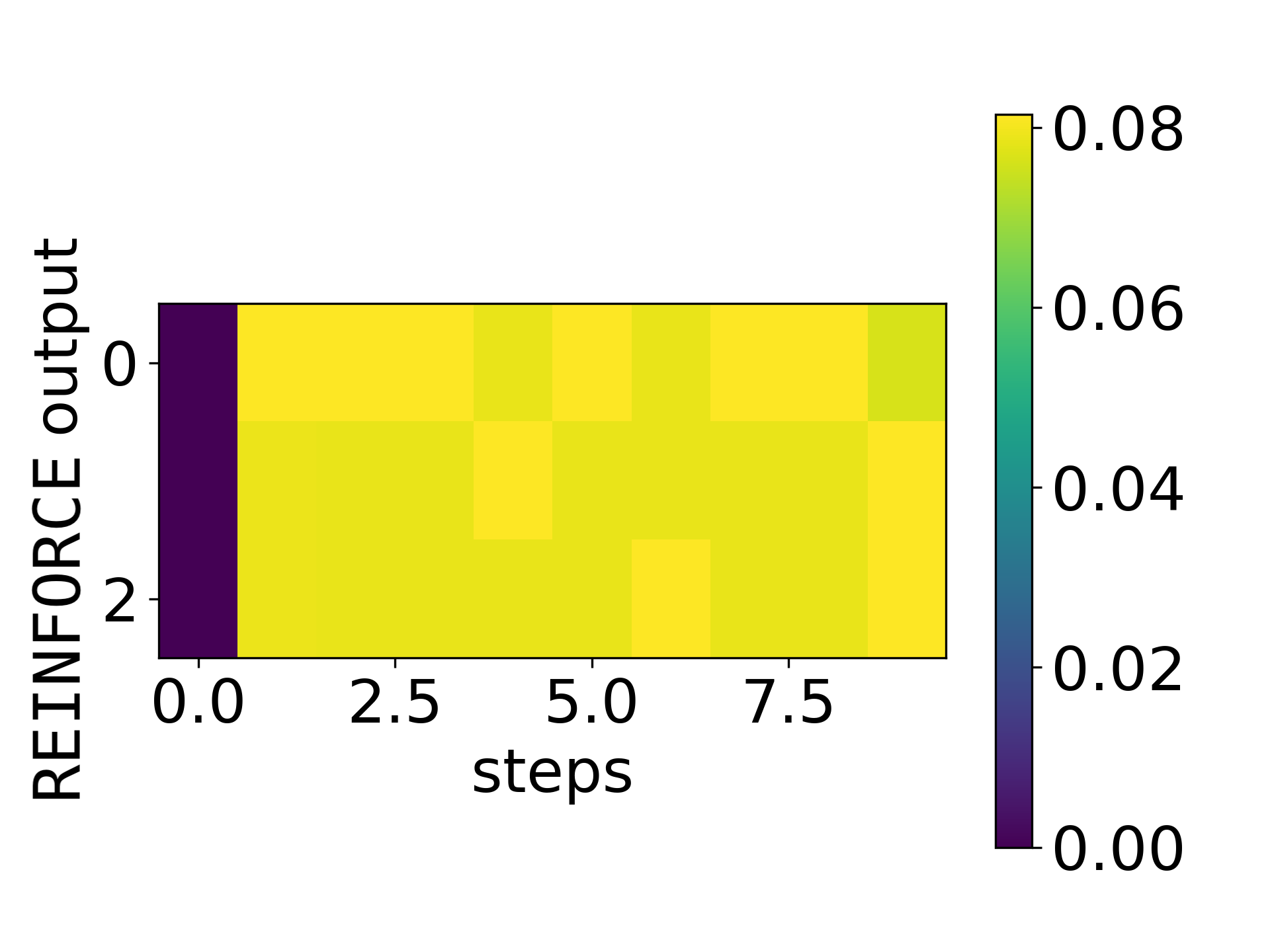}
 & 
    \includegraphics[width=0.3\textwidth]{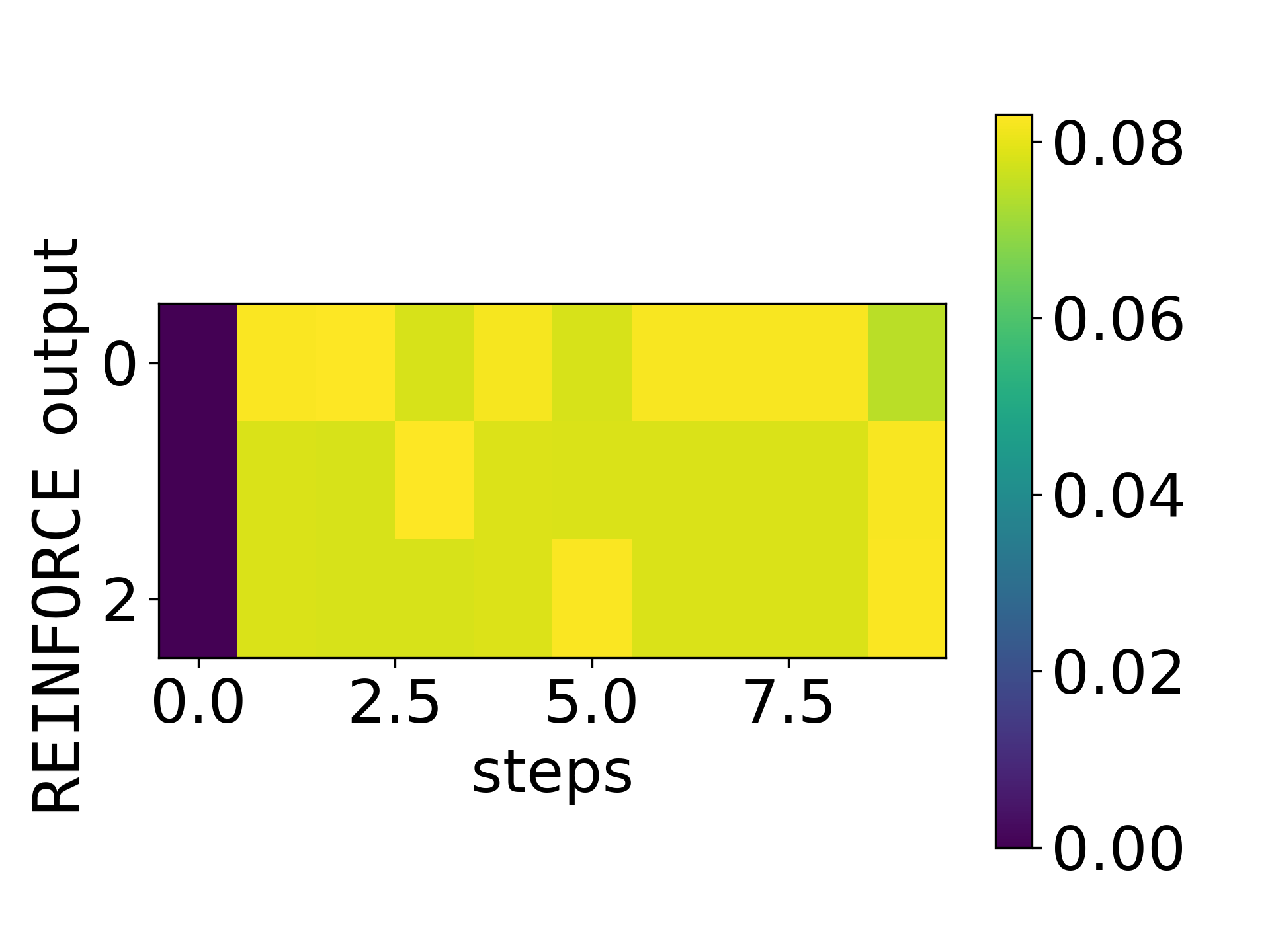}
 \\ 
    \includegraphics[width=0.3\textwidth]{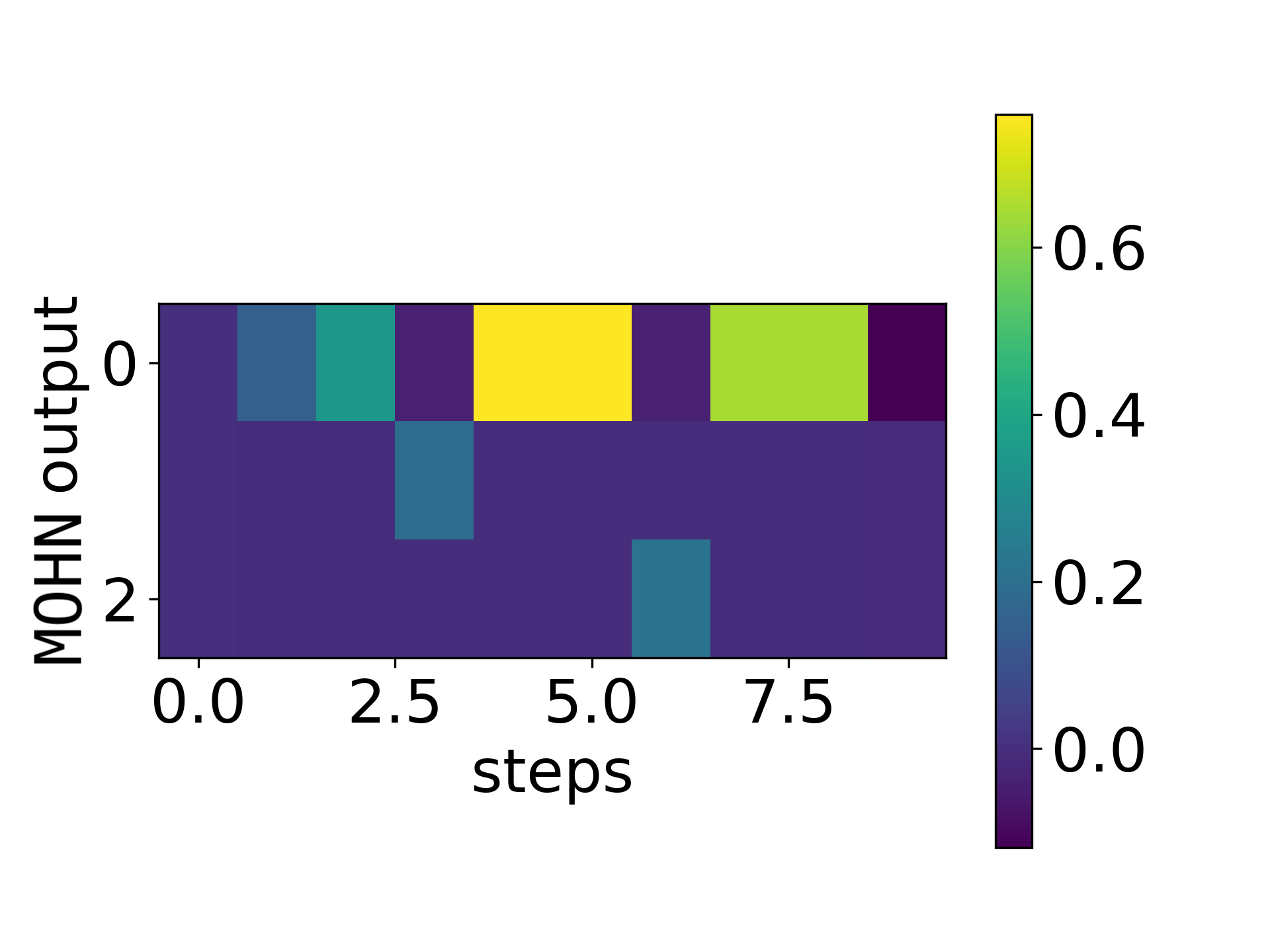}
 & 
    \includegraphics[width=0.3\textwidth]{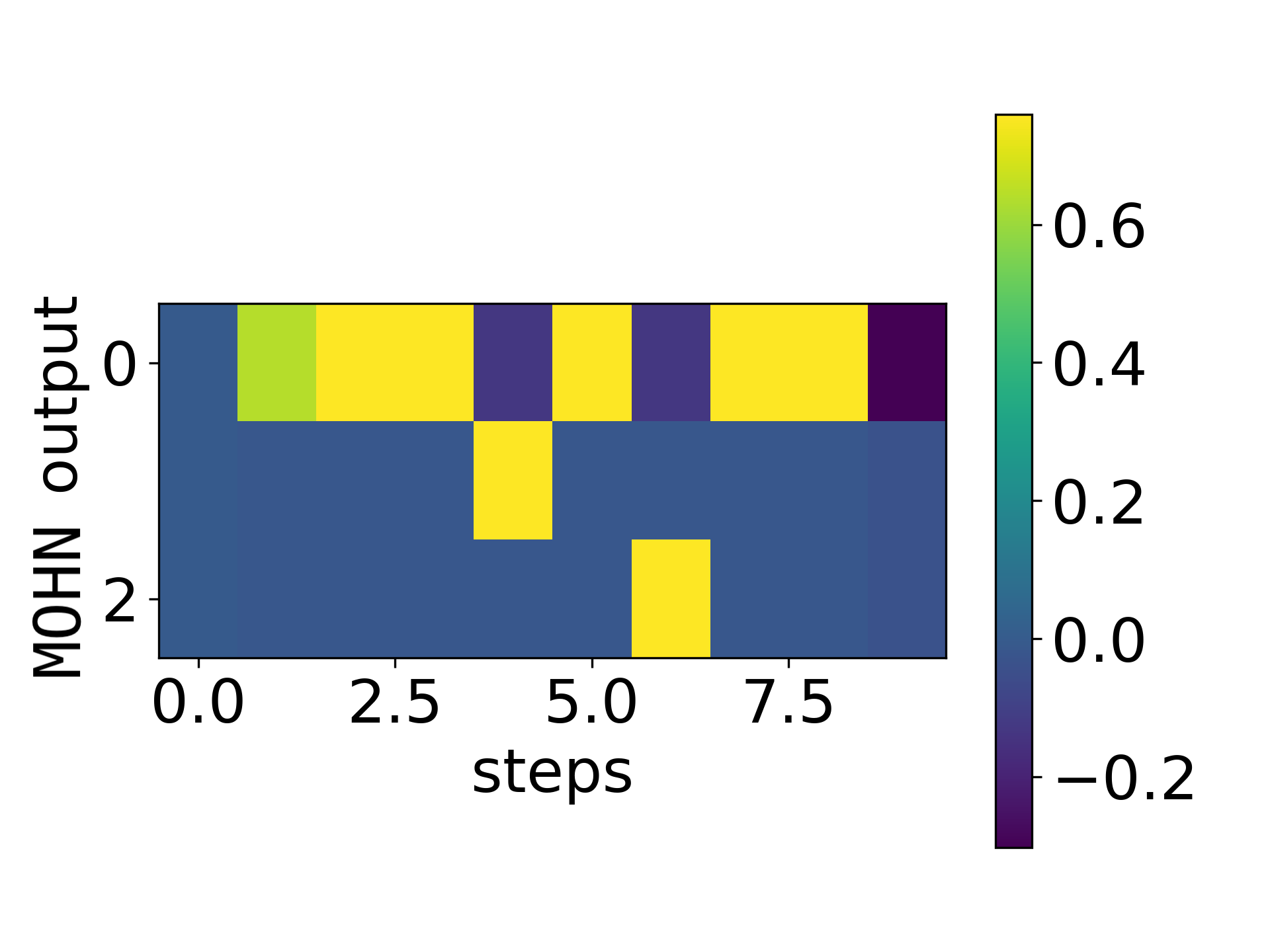}
 &  
    \includegraphics[width=0.3\textwidth]{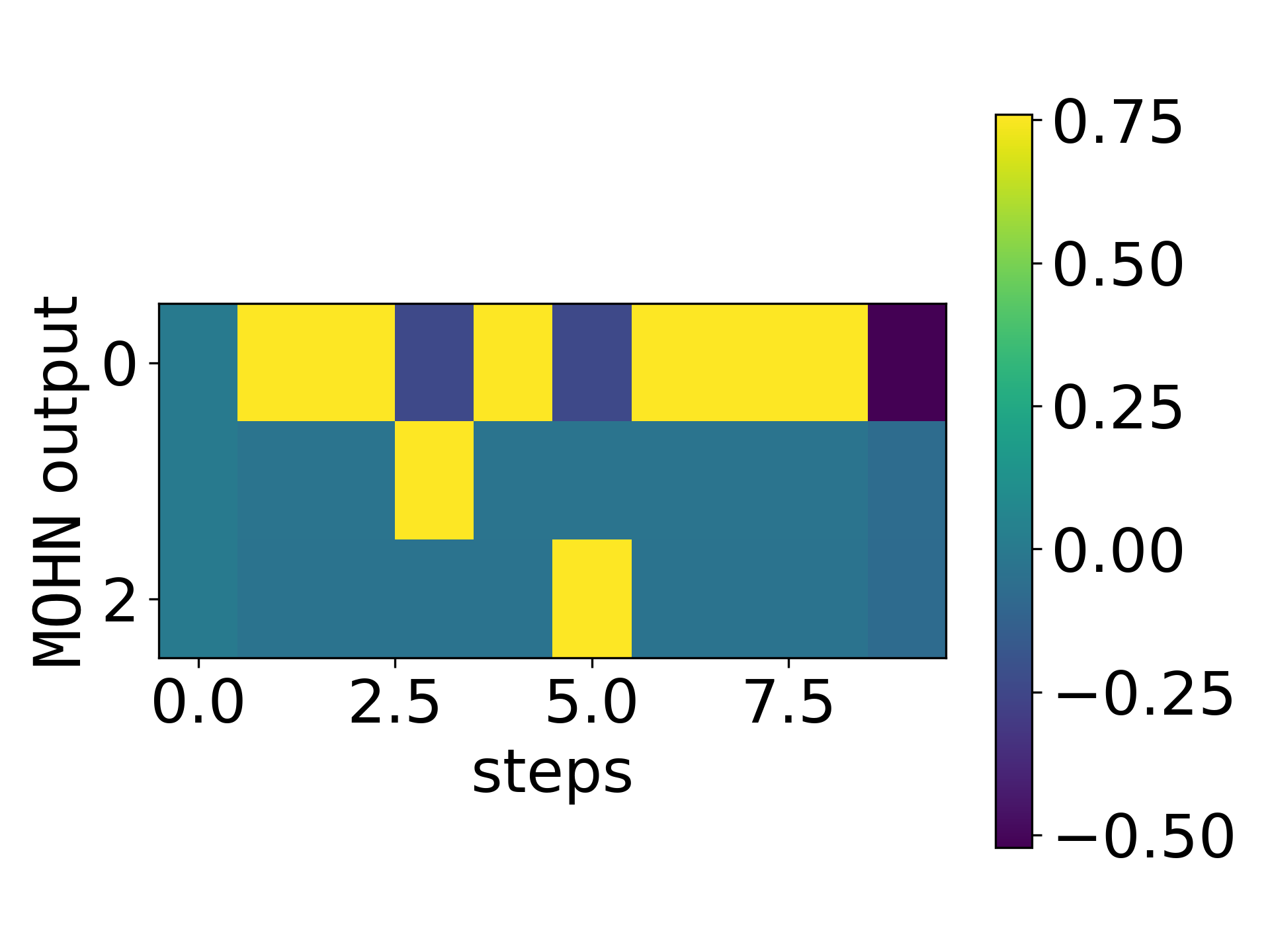}
 \\ 
\end{tabular}
\caption{Decision made by DQN (1st row), REINFORCE (2nd row) and MOHN (3rd row) at episode 2 (1st column), 5 (2nd column) and 10 (3rd column). The agents were guided to the goal manually for each episode to see how quickly they will learn the correct path. The correct sequence of actions is to take action 0 in wait state, action 1 in the first decision point and action 2 in a second decision point.  MOHN is the fastest, while DQN is struggling to reach good decisions in a few episodes.}
\label{fig:single_episode_no_noise}
\end{figure*}

\subsubsection{Importance of eligibility traces and high learning rates for quick learning} Fig.~\ref{fig:single_episode_no_noise} shows the decision made at each state at episodes 2, 5 and 10 for DQN, REINFORCE and MOHN. As can be seen, MOHN can learn from fewer scoring episodes than DQN and REINFORCE. This ability is vital in sparse reward problems as a reward is encountered rarely and needs to be utilized efficiently.
\begin{figure*}[htb]
\begin{tabular}{l|l|l}
Episode 2 & Episode 5 & Episode 10   \\ \hline
    \includegraphics[width=0.3\textwidth]{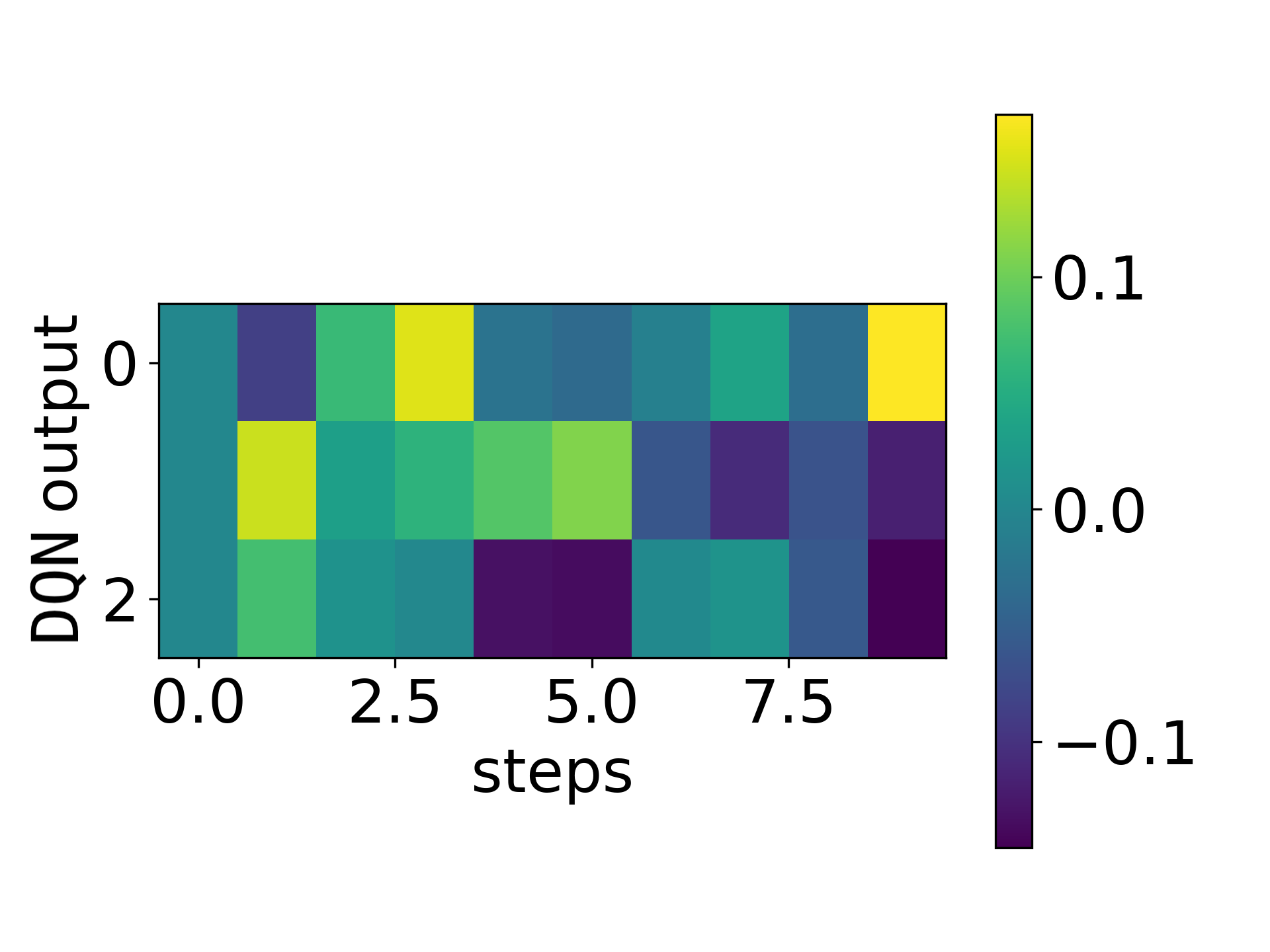}
 & 
    \includegraphics[width=0.3\textwidth]{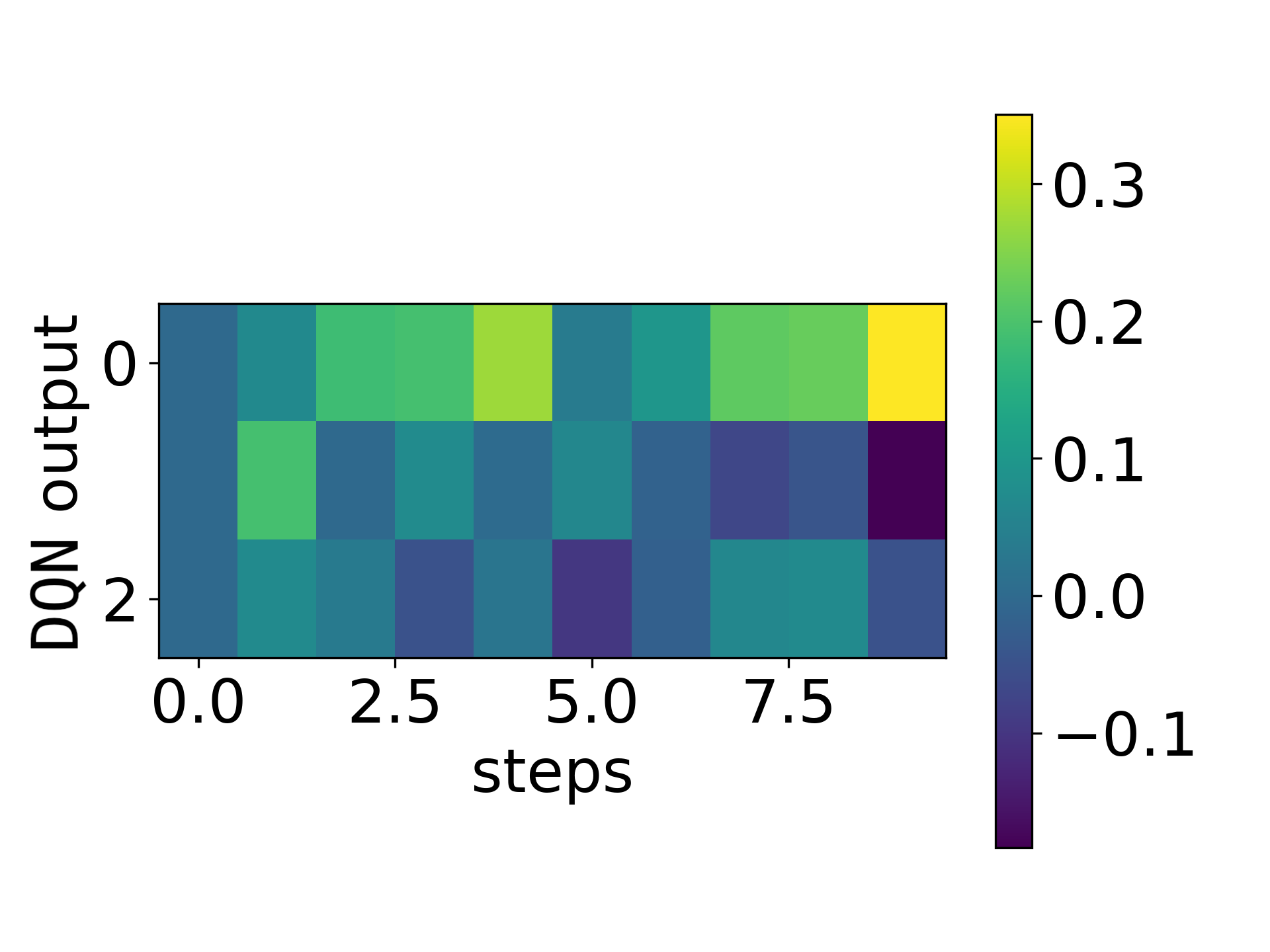}
 & 
    \includegraphics[width=0.3\textwidth]{out_single_episodes/dqn_out_single_episode_10.png}
 \\ 
    \includegraphics[width=0.3\textwidth]{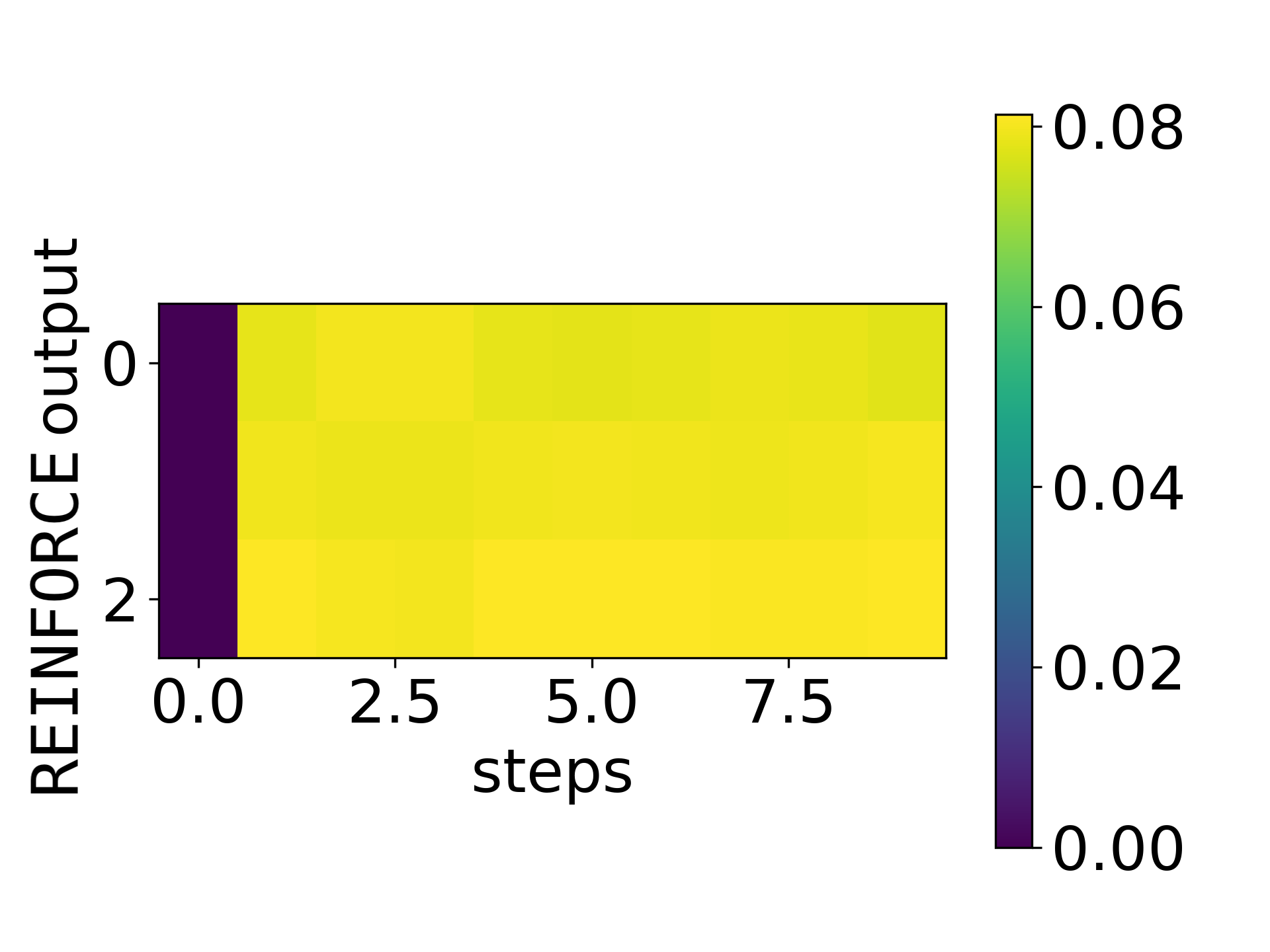}
 & 
    \includegraphics[width=0.3\textwidth]{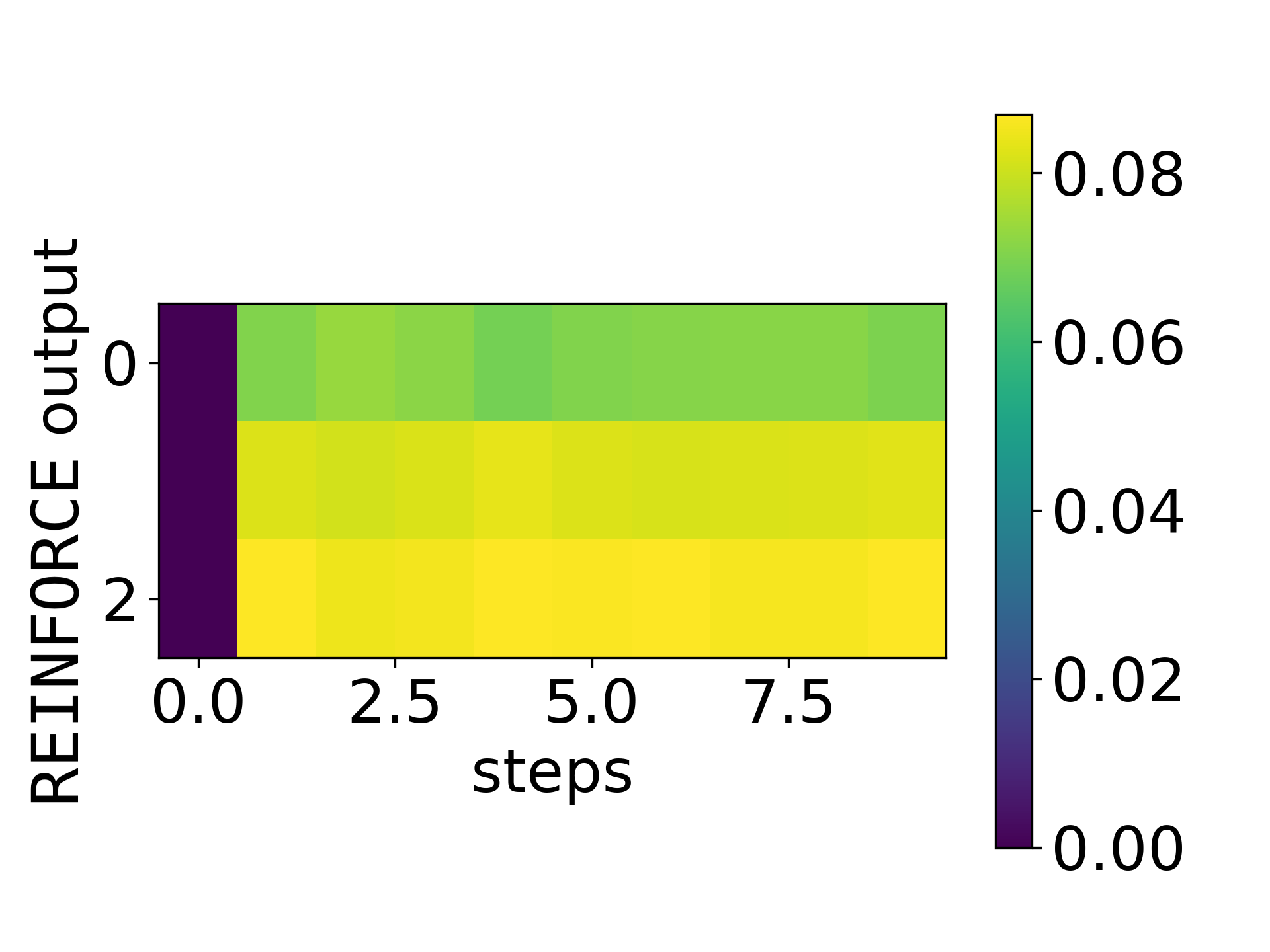}
 & 
    \includegraphics[width=0.3\textwidth]{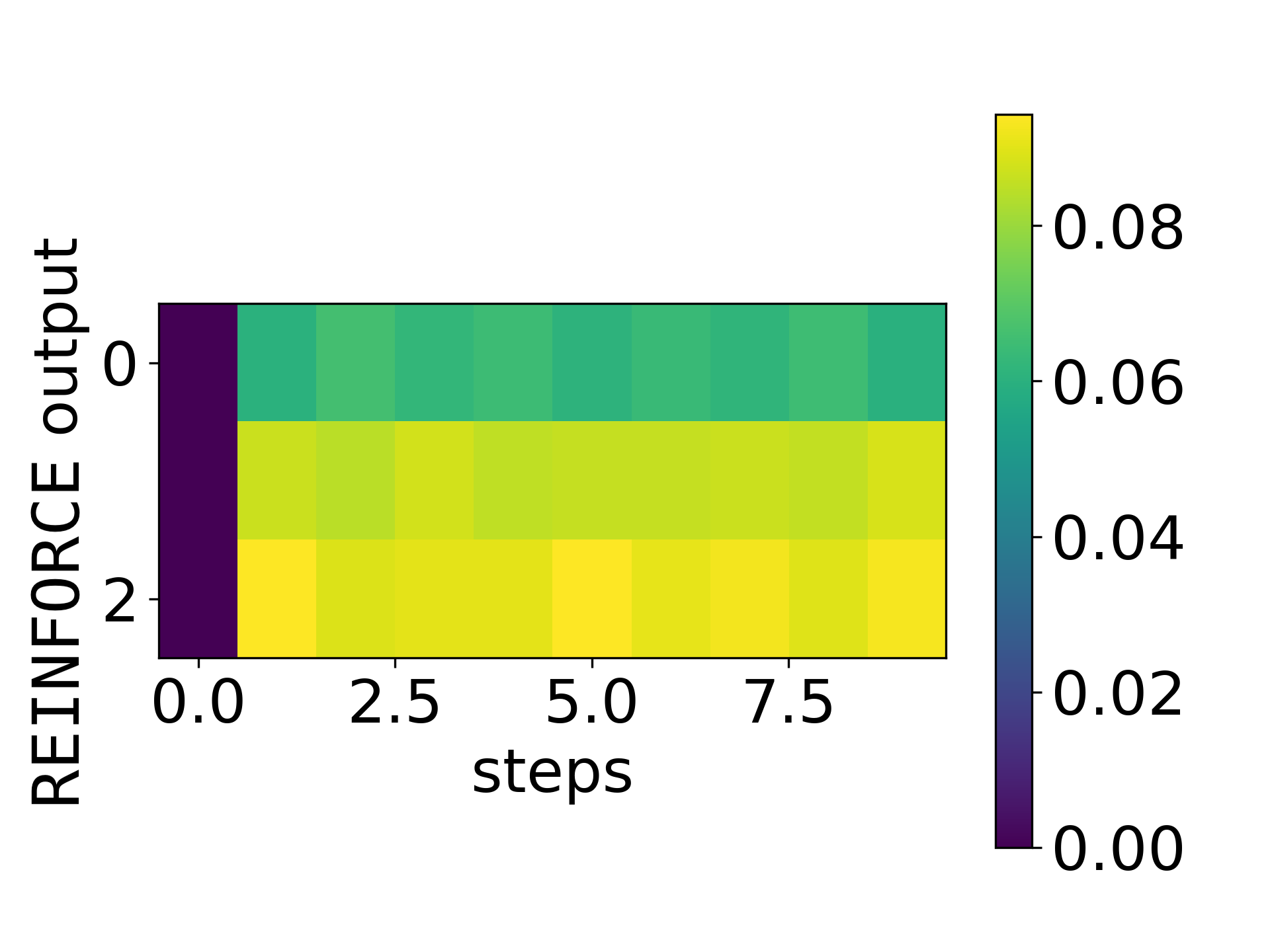}
 \\ 
    \includegraphics[width=0.3\textwidth]{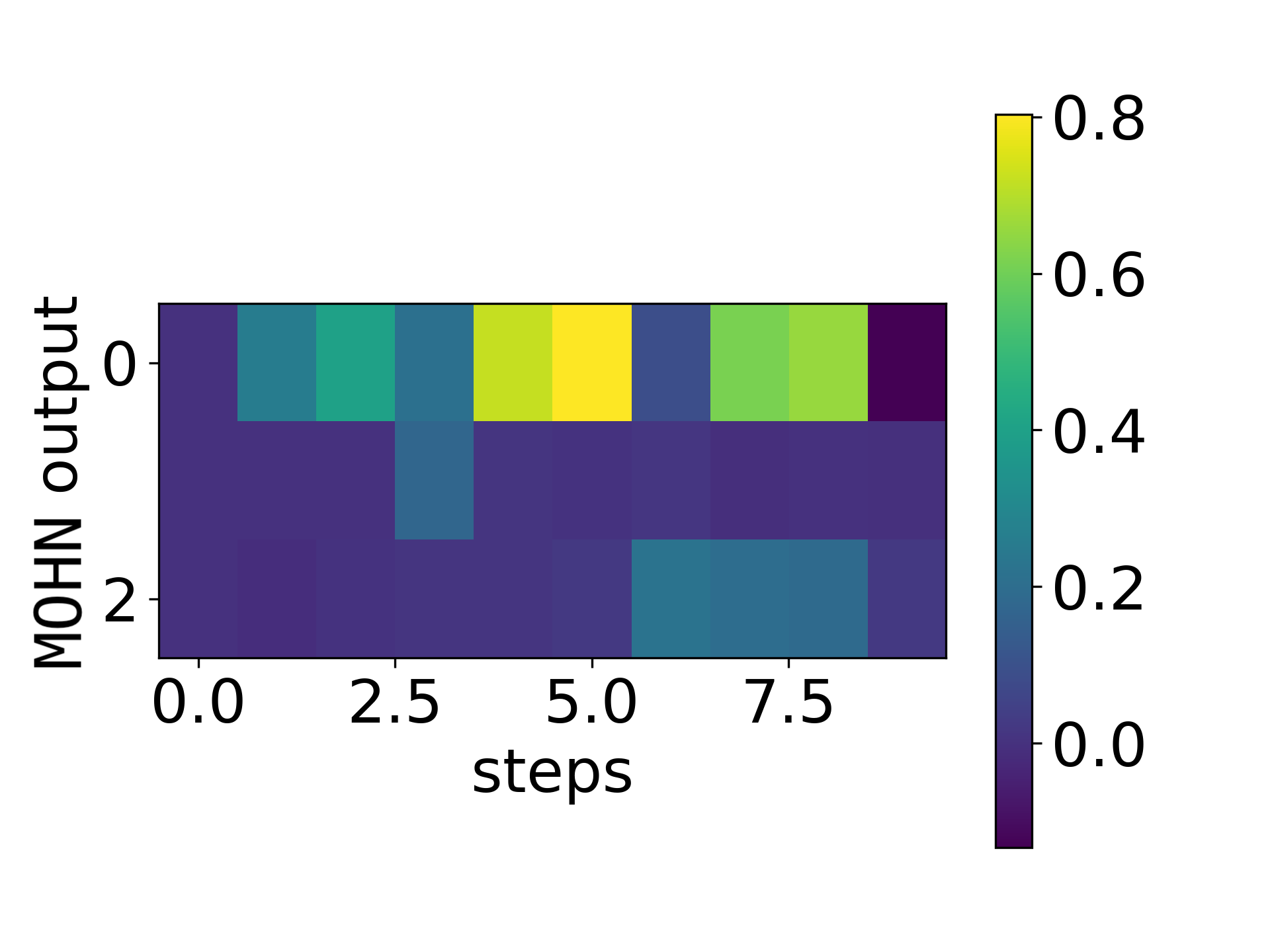}
 & 
    \includegraphics[width=0.3\textwidth]{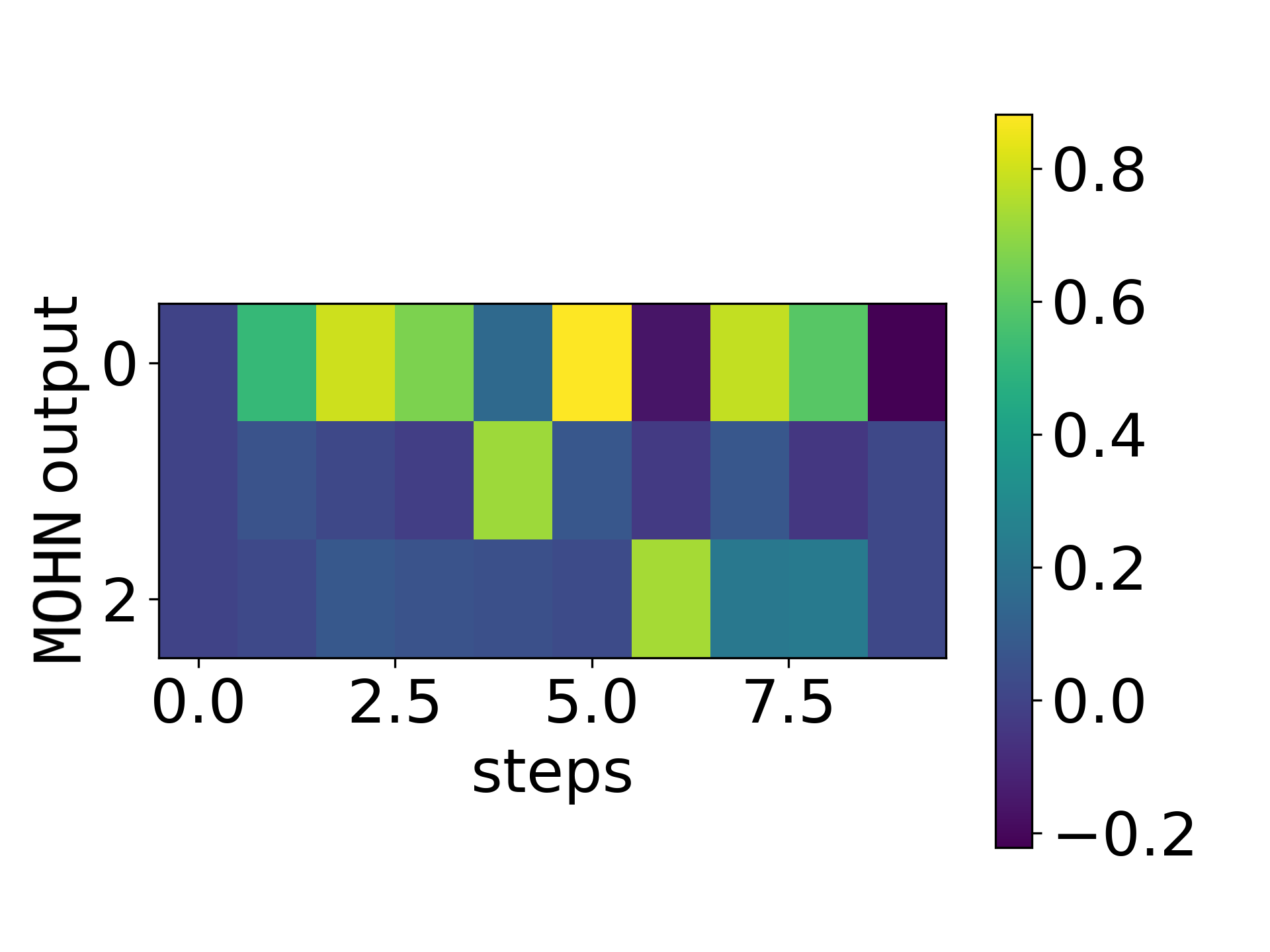}
 &  
    \includegraphics[width=0.3\textwidth]{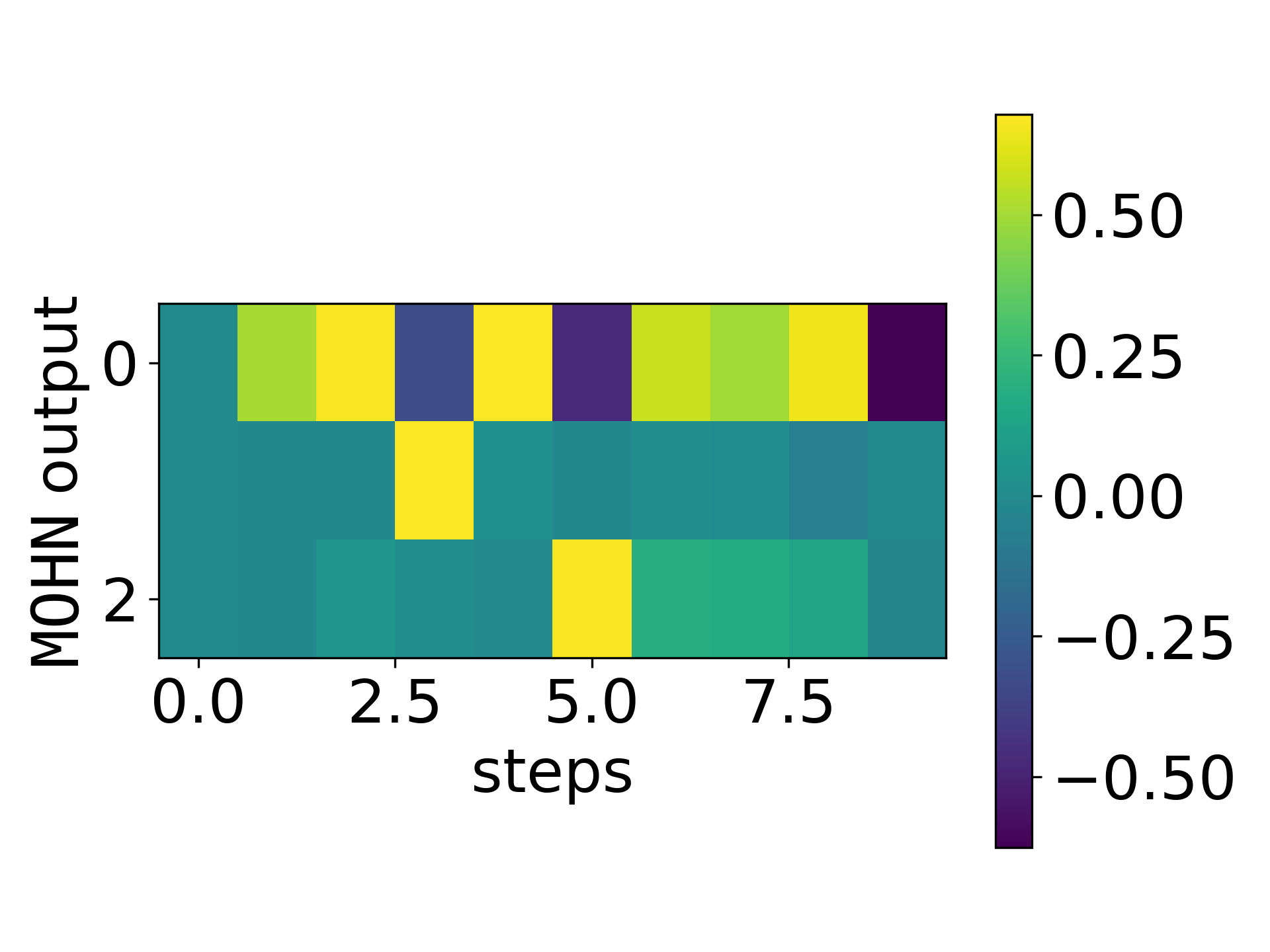}
 \\ 
\end{tabular}
\caption{Decisions made by DQN (1st row), REINFORCE (2nd row) and MOHN (3rd row) at episode 2 (1st column), 5 (2nd column) and 10 (3rd column) in a problem with noise. The agents were guided to the goal manually for each episode to see how quickly they learn the correct path. The correct sequence of actions is to take action 0 in wait state, action 1 in the first decision point and action 2 in a second decision point. MOHN is the fastest, while DQN and REINFORCE are struggling to reach good decisions in few episodes due to noise.}
\label{fig:single_episode_noise}
\end{figure*}

\subsubsection{Importance of rare correlations for coping with noisy inputs} Fig.~\ref{fig:single_episode_noise} shows the same comparison between three methods, this time with added noise 10\% to the inputs (see Appendix 3 for input space details). In this scenario, only MOHN was able to learn correct decisions in 10 episodes. This demonstrates another key mechanism of MOHN: utilization of rare correlations. This experiment shows that MOHN exploits rare correlations to make associations between key observations, actions and rewards.

\subsection{Analysis of feature learning.}
\begin{figure*}[h]
    \centering
    \includegraphics[width=0.8\textwidth]{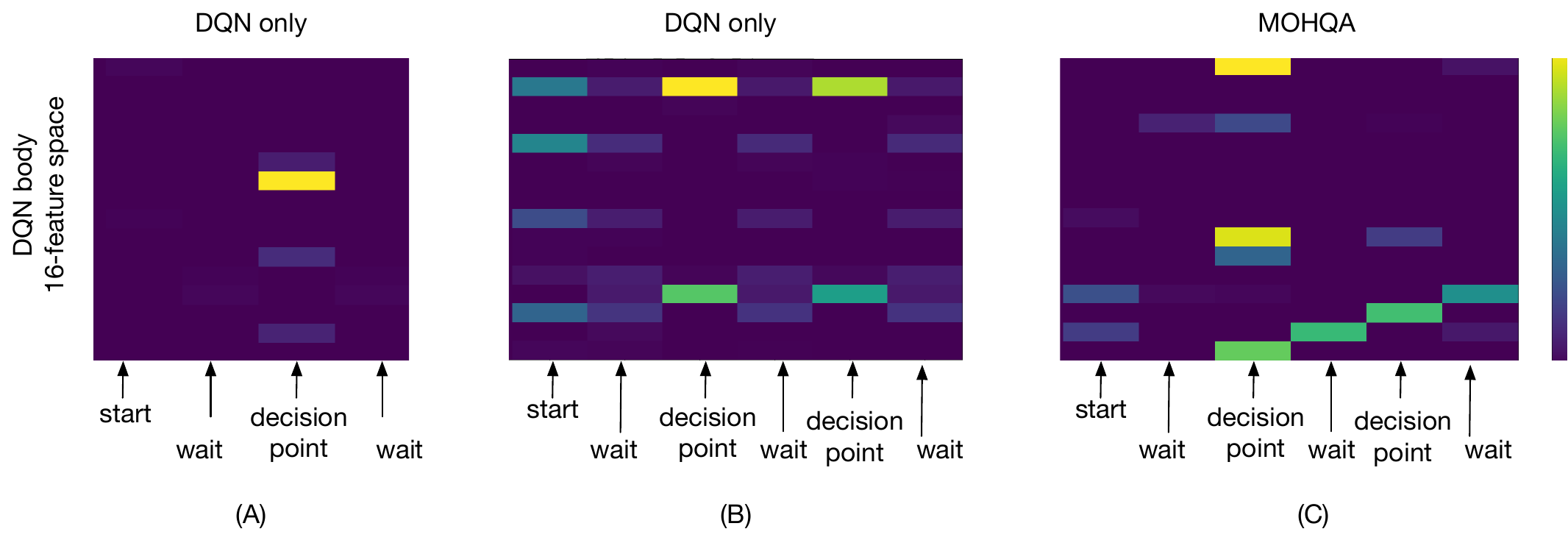}
    \caption{Features learned by the DQN without and with the MOHQA. (A) In a one-decision-point CT-graph, DQN correctly learns that the observations before and after the decision points are the same, so the TD error cannot be computed. (B) In a two decision-point CT-graph, DQN learns features that are the same for wait states and decision points: this allows navigation to a leaf node but random actions are taken at decision points. (C) The MOHQA helps with selecting the correct actions, and as a consequence, DQN learns different features for the two different decision points. }
    \label{fig:features}
\end{figure*}

To better understand the need for a new loss function $L(\Phi)$ (Eq. \ref{loss}) and how it affects feature learning ($v_{b}$ from Fig. \ref{fig:MOHQA}), three experiments are performed (Fig. \ref{fig:features}): (A) in a one-decision-point CT-graph, (B) in a two-decision-point CT-graph with confounding observations, both using the standard DQN's loss function, and finally (C) a two-decision-point CT-graph with confounding observations and the newly proposed loss function. Each of the snapshots of learned feature space were taken after the agent started consistently scoring maximum reward.

Fig. \ref{fig:features}(A) shows the learned feature space output throughout one episode. As expected, DQN learns similar high level features from different observations if those require the same action. Wait states that require the wait-action $a^0$ are distinguishable from decision points that require act-actions.

A similar situation is observed in a longer CT-graph with two decision points and confounding observations Fig. \ref{fig:features}(B). In this case, the two decision points had unique observations, thus making the problem observable, but only at decision points. DQN learns two distinct features spaces one for wait states and one for decision points. This is reasonable because these two state types require either action $a^0$ (wait state) or actions $a^1$ or $a^2$ in the decision point. However, the network is unable to distinguish between two decision points due to issues with propagating TD-error through confounding wait states. This confusion between the first and the second decision point highlights a problem DQN faces when trying to solve confounding POMDPs: if DQN cannot learn the path to the reward, it cannot also learn the separate features that would enable correct decisions.  

Finally, in the third experiment, same environment is used as in Fig. \ref{fig:features}(B), but the newly proposed loss function is utilized instead of the standard DQN's one. In Fig. \ref{fig:features}(C) the feature space clearly shows a difference between the first and second decision point. The MOHN, by suggesting optimal actions to the DQN, was able to also lead the DQN to learn different features for different decision points, which DQN alone could not achieve. In this last test, the output values reveal the inner working of the MOHQA architecture: DQN suggests the wait-action at wait states, and expresses equal preferences for both act actions $a^1$ and $a^2$. The MOHN contributes by biasing the decision towards the act-action (either $a^1$ or $a^2$) that is associated with the future reward.

\subsection{Computational Speed and Memory Comparison}
To analyse the computational effects of adding the MOHN to DQN, MOHQA and DQN are compared in terms of computational speed and GPU memory usage. Both DQN and MOHQA are run for 100,000 episodes on the same machine three times to minimize any variability in performance. The results of these runs are summarized in tables~\ref{tab:DQN_speed} and~\ref{tab:MOHQA_speed}.

It can be seen that MOHQA is about $66\%$ slower than the original DQN implementations. This is mostly due to operations in MOHN. In general backpropagation has been extensively studied and improved over time and it has efficient libraries written for it. On the other hand the learning mechanism in MOHN is not currently optimized. In particular, some operations such as finding maximum/minimum correlations are very computationally expensive. Comparing memory, a small increase of a GPU memory usage can be attributed to the extra layer and associated eligibility traces storage.

\begin{table}[htb!]
\begin{tabular}{|p{1.7cm}|p{1cm}|p{1.7cm}|p{1.7cm}|}
\hline
 Time (seconds)& Steps  & Time per step (milliseconds)  & GPU memory (per step) \\ \hline
 $419.51$ & $278477$ & $1.51$  & $971$ \\ \hline
 $459.85$ & $309650$ & $1.49$  & $970$  \\ \hline
 $463.97$ & $309135$ & $1.50$  & $971$ \\ \hline
\end{tabular}
\caption{Computational cost of DQN}
\label{tab:DQN_speed}
\end{table}

\begin{table}[htb!]
\begin{tabular}{|p{1.7cm}|p{1cm}|p{1.7cm}|p{1.7cm}|}
\hline
 Time (seconds)& Steps  & Time per step (milliseconds)  & GPU memory (per step) \\ \hline
 $669.86$ & $277243$ & $2.42$  & $1027$ \\ \hline
 $627.70$ & $257465$ & $2.44$  & $1027$  \\ \hline
 $645.41$ & $262397$ & $2.46$  & $1026$ \\ \hline
\end{tabular}
\caption{Computational cost of MOHQA}
\label{tab:MOHQA_speed}
\end{table}

\subsection{Comparisons in the CT-graph}
Simulations were performed for a range of CT-graphs and compared with TD-learning approach DQN; three memory based approaches: classical QRDQN+LSTM, Backpropamine and AMRL; policy based approach REINFORCE; and hybrid of policy and TD-learning A2C\footnote{The simulation code for MOHQA is available at https://github.com/pladosz/MOHQA.git.}. It is worth noting that AMRL is the most modern approach to compare against and it was developed to cope with similar type of POMDPs requiring memory. Backpropamine was implemented similarly to MOHN only on a final layer of the network. This was done for two reasons. First, memory modules are normally implemented only on a final layer\cite{hausknecht2015deep}. Second, in their original implementation\cite{miconi18a,Miconi2020} the test in reinforcement learning settings was done on the high-level features. Six simulations with different seeds were run and averaged for each algorithm on each of the CT-graph experiments. Memory in QRDQN+LSTM and Backpropamine was truncated at the end of each episode, while in AMRL truncation happened at the end of the iteration. We tuned all hyper parameters manually to the best performance of each baseline, for details regarding them see supplementary material 1. In general we split benchmarks in the CT-graph into two (rather arbitrary) categories: simple and complex, where simple are ones which most approaches can solve, while complex are ones where algorithms start to struggle.

\subsubsection{Simple CT-graph problems}

\begin{figure*}
    \centering
    \begin{subfigure}[t]{0.32\textwidth}
        \centering
  \includegraphics[width=\textwidth]{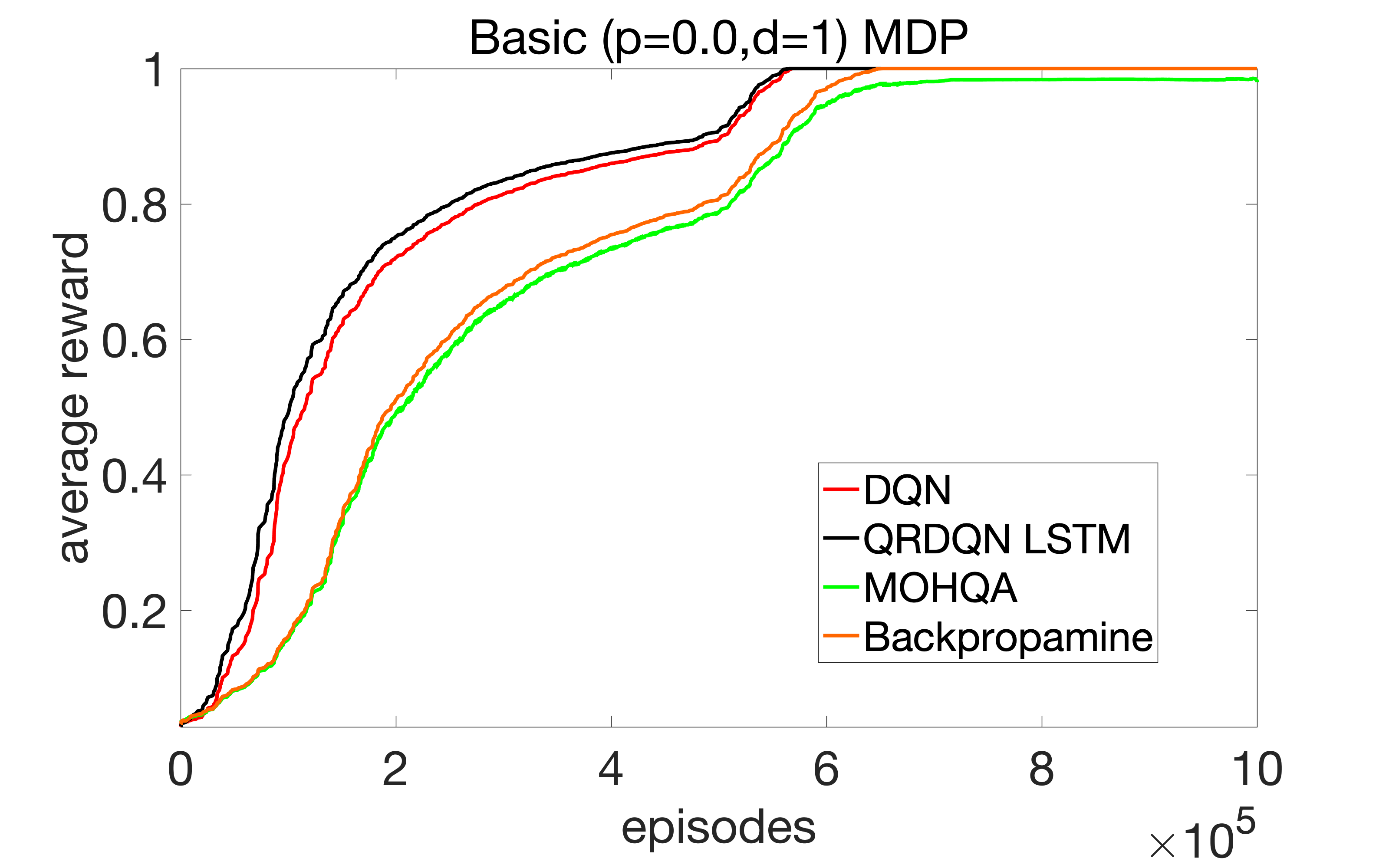}
  \caption{}
\label{fig:single_maze_working}
    \end{subfigure}%
    \begin{subfigure}[t]{0.32\textwidth}
        \centering
    \includegraphics[width=\textwidth]{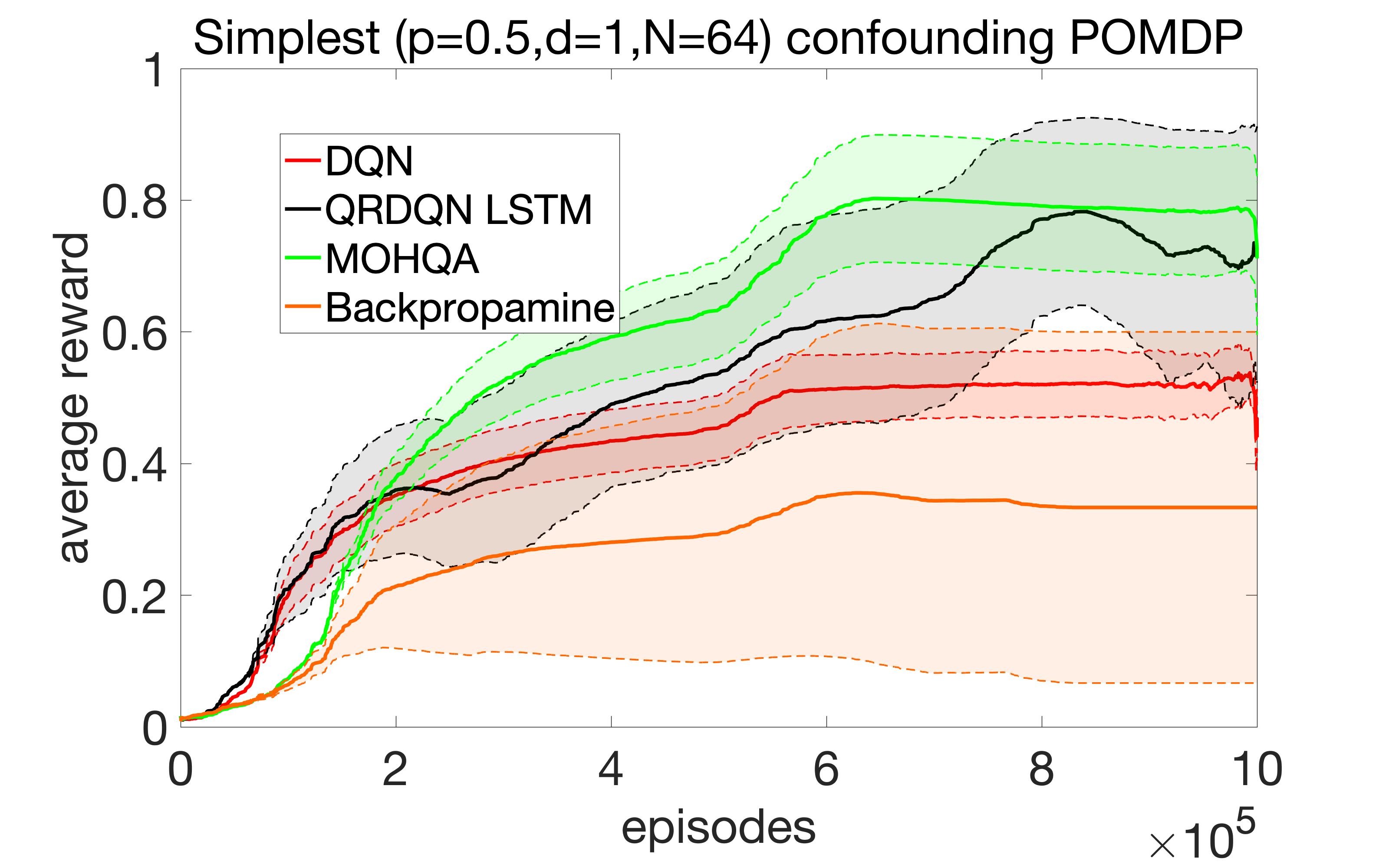}
   \caption{}
 
\label{fig:single_maze_broken}
    \end{subfigure}
    \begin{subfigure}[t]{0.32\textwidth}
        \centering
  \includegraphics[width=\textwidth]{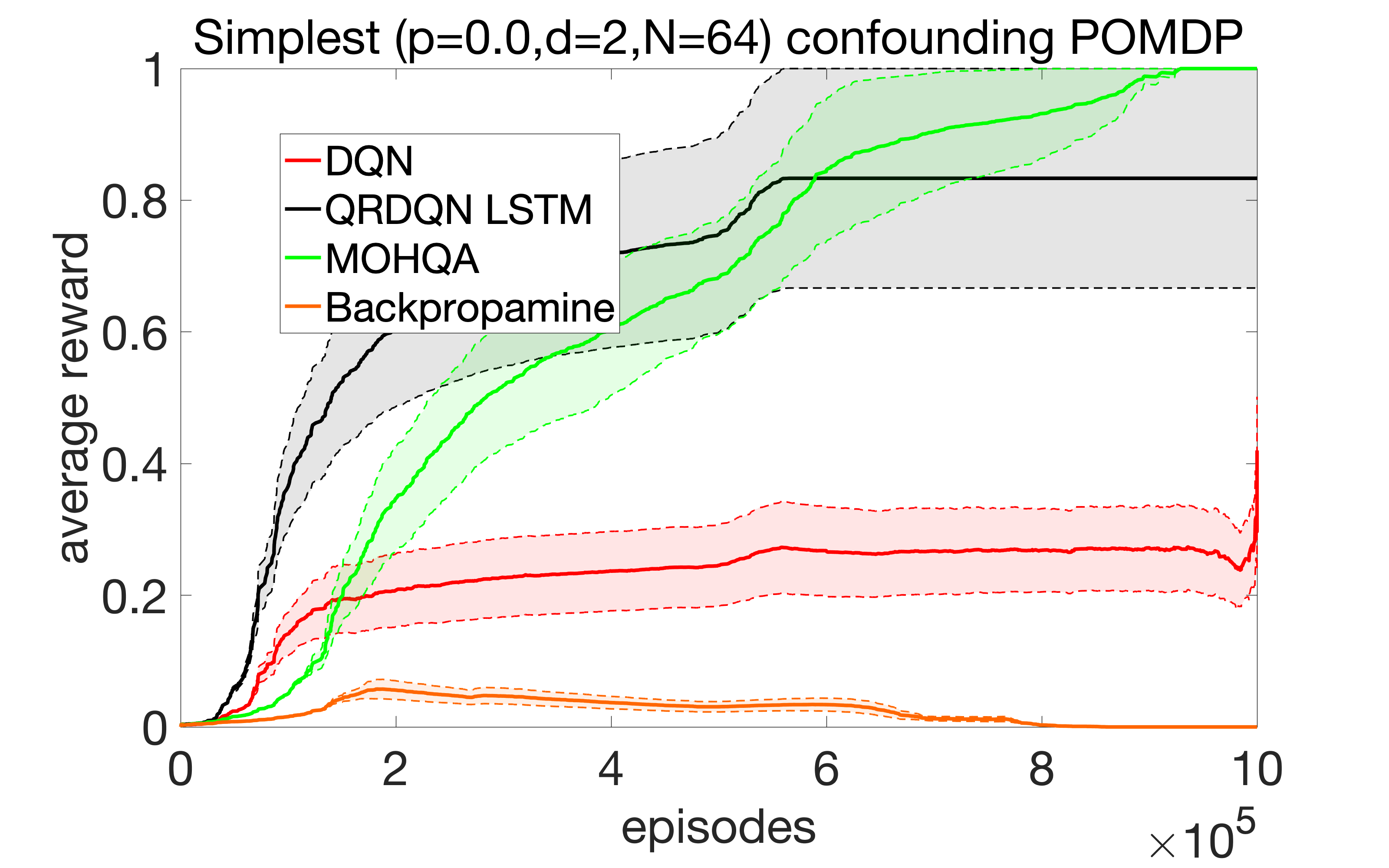}
  
\label{fig:double_maze_working}
\caption{}
    \end{subfigure}
    \caption{Performance on the simpler versions of the CT-graph. (a) In an MDP benchmark test (each observation is unique, and therefore reveals a unique state, $p=0$): as expected, all algorithms learn the optimal behavior. (b) POMDP ($p=0.5$, $N=64$): wait states provide similar observations: DQN starts to struggle and collects 50\% of the reward. The addition of the MOHN head means DQN is able to solve the problem better. (c) With a delay probability ($p=0$), ($d=2$) and confounding observations ($N=64$), MOHQA and QRDQN+LSTM appears to learn a a good strategy, while DQN performance drops.}
    \label{fig:single_maze}
\end{figure*}
We run the the simplest MDP version of the one-decision-point CT-graph (Fig.~\ref{fig:single_maze}(a)) to verify the correctness of implementations. In Fig.\ \ref{fig:single_maze}(b), the confounding observations are introduced to the CT-graph by removing uniqueness from the wait states. MOHQA and other approaches are able to solve the problem while DQN starts to struggle as it cannot learn correct actions. Fig.~\ref{fig:single_maze}(c) shows the results on the simplest of two decision point CT-graphs with delay probability of ($p=0$) and depth of ($d=2$).  This figure shows further advantage of MOHQA over DQN. Note that the history of observations is still relatively simple as histories repeat quite often, thus the good performance of QRDQN+LSTMs (for analysis of histories of observations see supplementary material 2).
\subsubsection{The Complex CT-graph}
\begin{figure*}[!htb]
    \centering
    \begin{subfigure}[t]{0.31\textwidth}
        \centering
        \includegraphics[width=\textwidth]{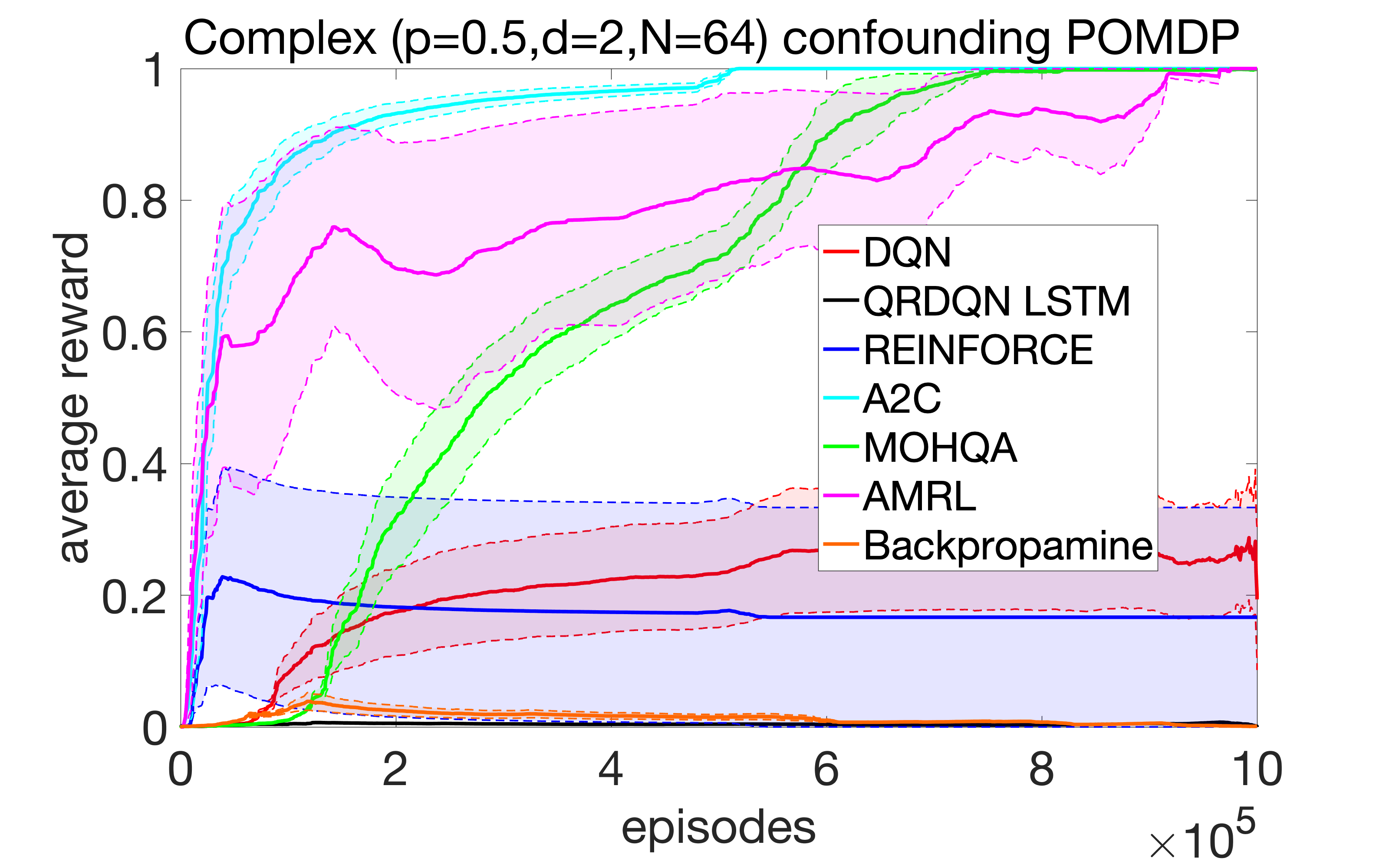}
        \label{fig:complex_1}
        \caption{}
    \end{subfigure}
    \begin{subfigure}[t]{0.31\textwidth}
        \centering
        \includegraphics[width=\textwidth]{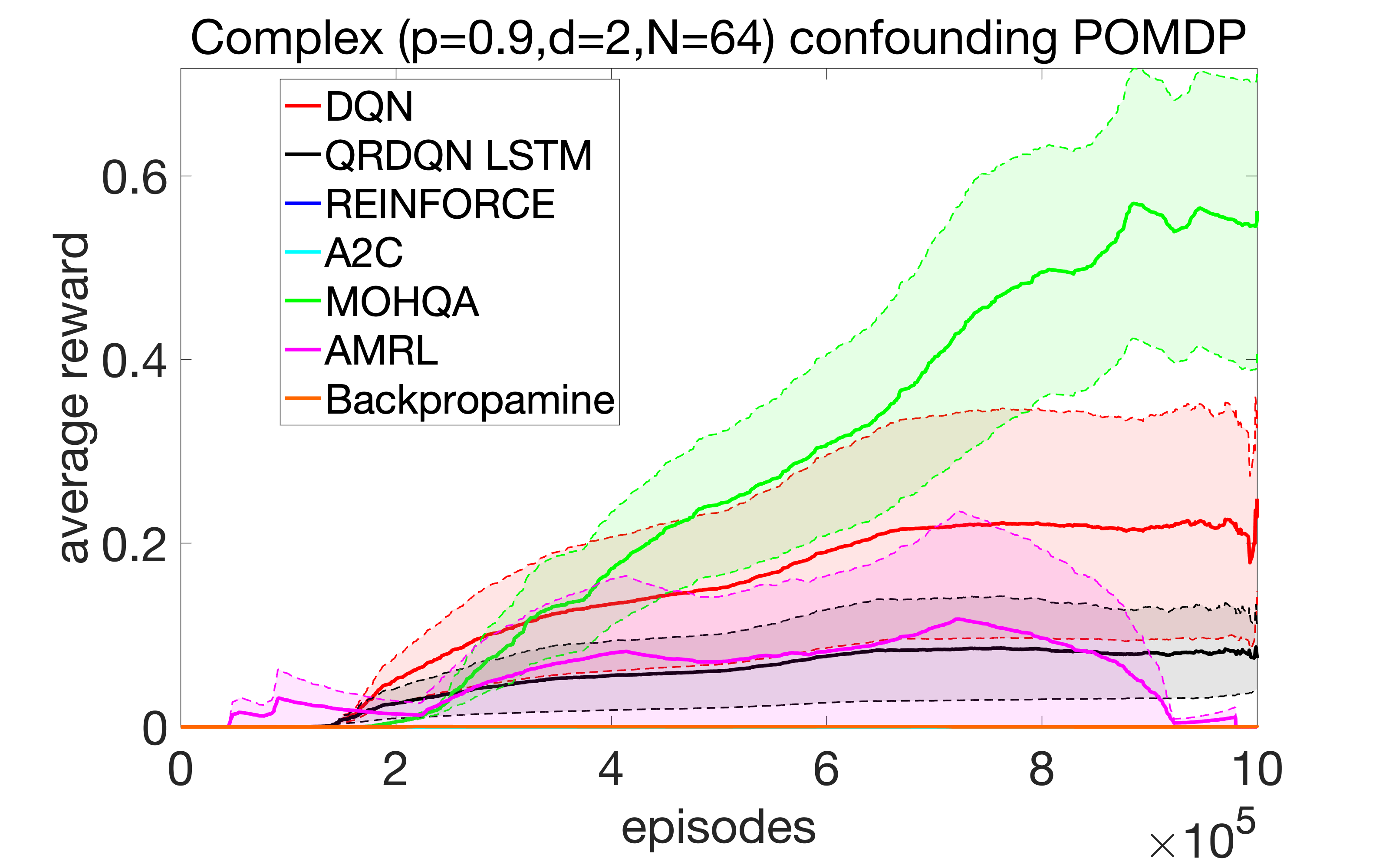}
        \label{fig:complex_2}
        \caption{}
    \end{subfigure}
    \begin{subfigure}[t]{0.31\textwidth}
        \centering
        \includegraphics[width=\textwidth]{pa_gs_2_d_2_bf_2_DelayProb_0_9_more_observations.png}
        \label{fig:complex_3}
        \caption{}
    \end{subfigure}
    \caption{Performance on a more complex CT-graph. As simpler baselines' performance has started deteriorating, A2C, REINFORCE and AMRL algorithms are added for extra comparison. (a) The problem with increased delay probability ($p=0.5$), depth of $d=2$ and confounding observations $N=64$ shows extremely fast convergence of A2C, and a reduction of performance of DQN, QRDQN+LSTM and Backpropamine. QRDQN+LSTM and Backpropamine might be struggling due to non-repeating histories, while DQN might suffer from TD-error propagation problems. Albeit a bit slower to converge, MOHQA is able to achieve the same performance as A2C. (b) The CT-graph with delay probability of $p=0.9$, depth of $d=2$ and confounding state number $N=64$ is an extremely sparse reward problem with confounding stimuli. On this problem MOHQA performs on par with AMRL, while beating other approaches. c) The CT-graph with delay probability of $p=0.9$, depth of $d=2$ and confounding observations $N=500$ shows MOHQA outperforms all of the other approaches in very confounding POMDPs.}
    \label{fig:double_maze}
\end{figure*}
For the complex CT-graph problems, we compared MOHQA against A2C, REINFORCE and AMRL. All tests are performed with depth ($d=2$) and three different configurations: i) $p=0.5$, $N=64$ (Fig.~\ref{fig:double_maze}(a)), ii) $p=0.9$, $N=64$ (Fig.~\ref{fig:double_maze}(b)) and iii) $p=0.9$, $N=500$ (Fig.~\ref{fig:double_maze}(c)), where $N$ is the cardinality of the set of confounding observations and $p$ is the delay probability. The two-decision-point CT-graph (Fig.~\ref{fig:double_maze}(a)) shows very similar trends to the simple CT-graph problems, with MOHQA still improving on performance of DQN, achieving similar performance to memory based AMRL. QRDQN+LSTM starts to struggle due to significantly increased number of possible histories of observations. Fig.~\ref{fig:double_maze}(b) shows deterioration of performance of A2C, QRDQN+LSTM, REINFORCE and DQN, to the point where MOHQA and AMRL show significant advantage of performance. Finally, in the most complex CT-graph (Fig.~\ref{fig:double_maze}(c)), MOHQA outperforms all other tested algorithms. Wide confidence intervals are a consequence of the 'all or nothing' reward structure. This means that agents either tend to know how to solve it and score 1 or they don't know and they score zero. Thus, a run that scored optimally in five out of six seeds reports an average normalised fitness of 0.8 with a standard deviation of +/- 0.14.

\subsection{Comparison in Malmo Benchmark}
\begin{figure}
   \centering
    \includegraphics[width=0.4\textwidth]{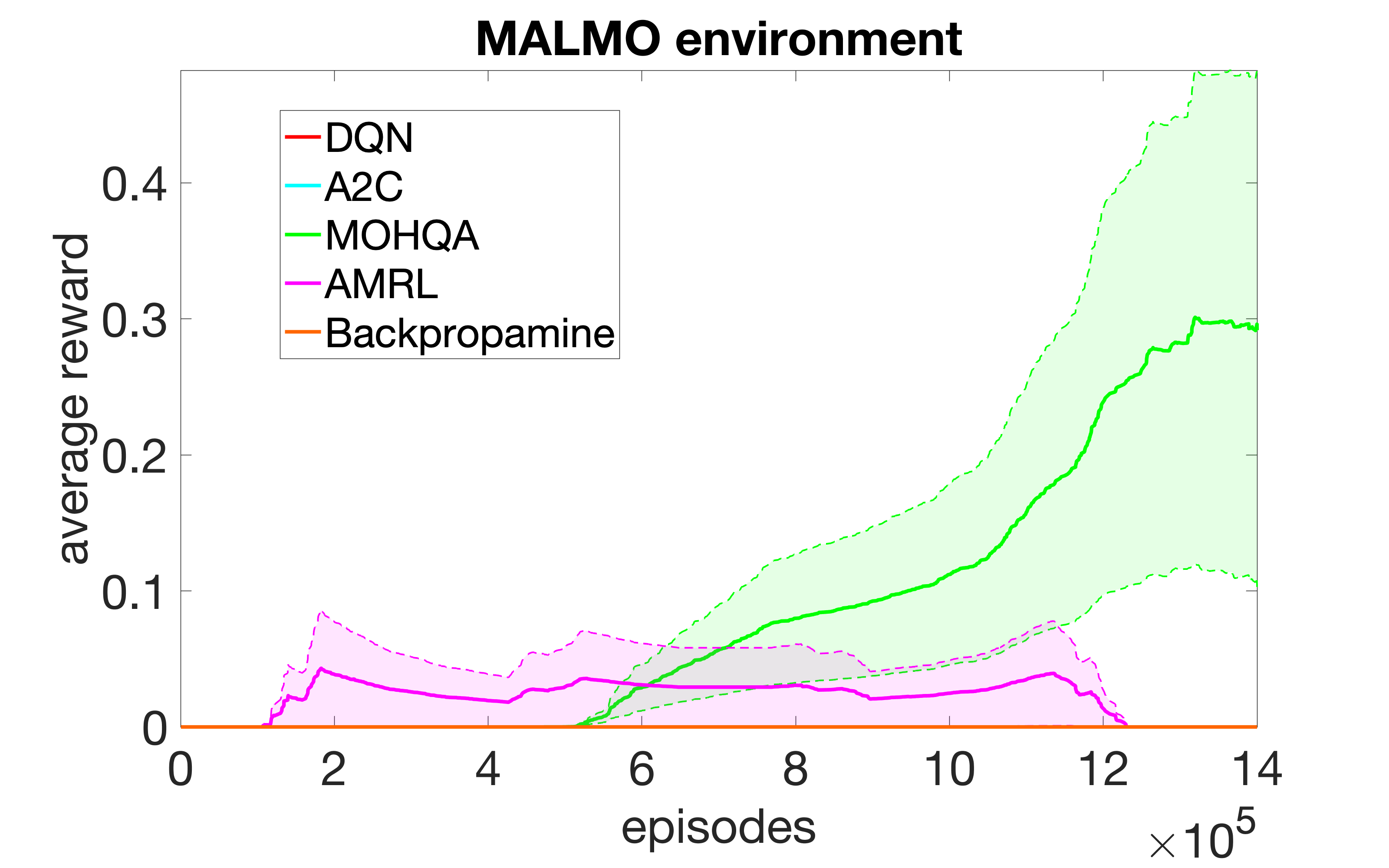}
   \caption{Comparison of results in Malmo benchmark. MOHQA is outperforming AMRL, DQN, A2C and Backpropamine.}
\label{fig:minecraft_results}
\end{figure}

The Malmo benchmark was configured as a maze with two decision points (see Fig.~\ref{fig:minecraft_overview}). An agent requires 30 steps to get from the home location to the end of the maze. Six simulations with different seeds were run and averaged. Fig.~\ref{fig:minecraft_results} shows that the MOHQA was able to achieve a higher average score than all other baselines. The result shows that the advantage achieved by MOHQA in the previous benchmarks is transferable to different domains.

\section{Discussion}

The confounding POMDP has proven to be a difficult problem for DQN, QRDQN+LSTM, REINFORCE, A2C and AMRL. The addition of the MOHN module to the standard DQN allowed the newly proposed MOHQA to achieve higher average scores than standard DQN. Such an advantage is achieved by the ability of MOHN to distinguish important decision states from confounding wait states and bridge the temporal gap between state-action pairs and rewards. 

REINFORCE and MOHQA appear to have very similar weight update scheme, yet MOHQA shows very clear advantage over REINFORCE. We hypothesise that this is a result of a more effective extraction of cause-effect links with delayed rewards because we use non-symmetrical Hebbian updates and rare correlated traces, effectively generating hypotheses about causal temporal links in the complex POMDPs dynamics, while ignoring inherent noise in the feature space.

The ability to find important decision points and bridge state-action-reward gaps was achieved using three key mechanisms: i) a novel implementation of STDP-inspired eligibility traces, ii) a novel use of modulated Hebbian learning in a deep RL and iii) a new deep RL architecture that integrates DQN with a Hebbian-based structure. STDP-inspired eligibility traces bridge state-action-reward gaps, allowing MOHQA to solve problems where TD-learning fails. Modulated Hebbian learning (Eq. 2) allows large weight updates without gradient explosions. This means learning happens quickly just from few scoring examples, which is particularly useful when the agent finds a reward rarely. The new deep RL architecture is used to guide DQN to learn to follow a good policy so that it provides good features to distinguish states. Each of the mechanisms is vital in achieving good performance.

An interesting consideration is that the MOHN also appears to be instrumental for guiding DQN to learn useful features. Due to the confounding observations that are provided during wait states, the baseline algorithms struggle to learn useful features of the decision points, which are instrumental to inform an optimal policy. Thus, learning the appropriate features depends on the actions which in turn depends on the features. While this chicken and egg problem is typical in deep reinforcement learning, the proposed MOHQA appears to facilitate the process of guiding the learning of useful features by discovering cause-effect relationships and offering guidance to the DQN underlying architecture. 
 
It is worth noting that memory does not help to solve problems in which the history of observations does not repeat (see performance of QRDQN+LSTM and AMRL). Firstly, recall that just like in real life, the exact history repeats itself very rarely in the CT-graph (see supplementary material 2 for detailed analysis). Secondly, the CT-graph and Malmo have significant gaps between key states and rewards at all times. Coping with those two problems require a significant number of samples, which very quickly become computationally infeasible. Even in memory approaches  designed to find key states such as FRMQN~\cite{Oh2016}, key states are learned on small problem sizes.

MOHQA is a first proof-of-concept and was tested on a limited set of sparse reward problems and compared with a limited number of benchmark algorithms. Further tests with other sparse reward problems, e.g., the Morris water maze and some ATARI games, will be essential to test the full potential of the approach. Yet, the present work suggests that a fundamentally simple confounding POMDP casts insights on the challenging problem of learning simultaneously a feature space and a policy.  

The proposed architecture, while proving  effective and posing a new learning paradigm, has some limitations. The MOHQA is more complex than a standard DQN network, and requires tuning of additional hyper parameters. However, to the best of the authors' knowledge, this is the first successful attempt to combine a modulated Hebbian network with a DQN network in a new RL architecture. An interesting future research direction is to implement MOHQA with different deep reinforcement learning algorithms such as A2C. This should extend the capabilities and improve stability to allow solving even more complex confounding POMDPs.

 MOHQA could also be used to help solve other problems such as autonomous driving and robotic delivery. In autonomous driving, roads have different lengths between junctions. Also, they are full of irrelevant features (cars, pedestrians, different building etc.) and the key decisions happen at the junctions. The reward is only given after reaching its destination. This is similar to the type of problems MOHQA has been able to solve. Other robotics tasks in stimulus-rich environments, such as tasks in houses or public places, are also full of confounding and changeable observations that are problematic for RL algorithms that have been tested only in video-game environments.  In this paper, we suggest and demonstrate that better algorithms are required to make RL more applicable to real-world problems.

\section{Conclusion}
This paper considers solving confounding POMDPs using a new neural architecture (MOHQA) for deep reinforcement learning. The key novelty in MOHQA is the addition of a modulated Hebbian learning network with neural eligibility traces (MOHN) to a standard DQN architecture. The objective is to provide basic RL algorithms with the ability to ignore confounding observations and delays, and associate key cause-effect relationships to delayed rewards. It was shown that the combination of DQN and MOHN can match and even outperform more advanced algorithms such as A2C, AMRL and QRDQN+LSTM on confounding POMDPs. While this is the first proof-of-concept study to propose a combined Hebbian and backpropagation-learned architecture for deep reinforcement learning, the promising results encourage further tests on a wider range of standard deep RL benchmarks.

\bibliographystyle{IEEEtran}
\bibliography{bibliography.bib}

\begin{IEEEbiography}[{\includegraphics[width=1in,height=1.25in,clip,keepaspectratio]{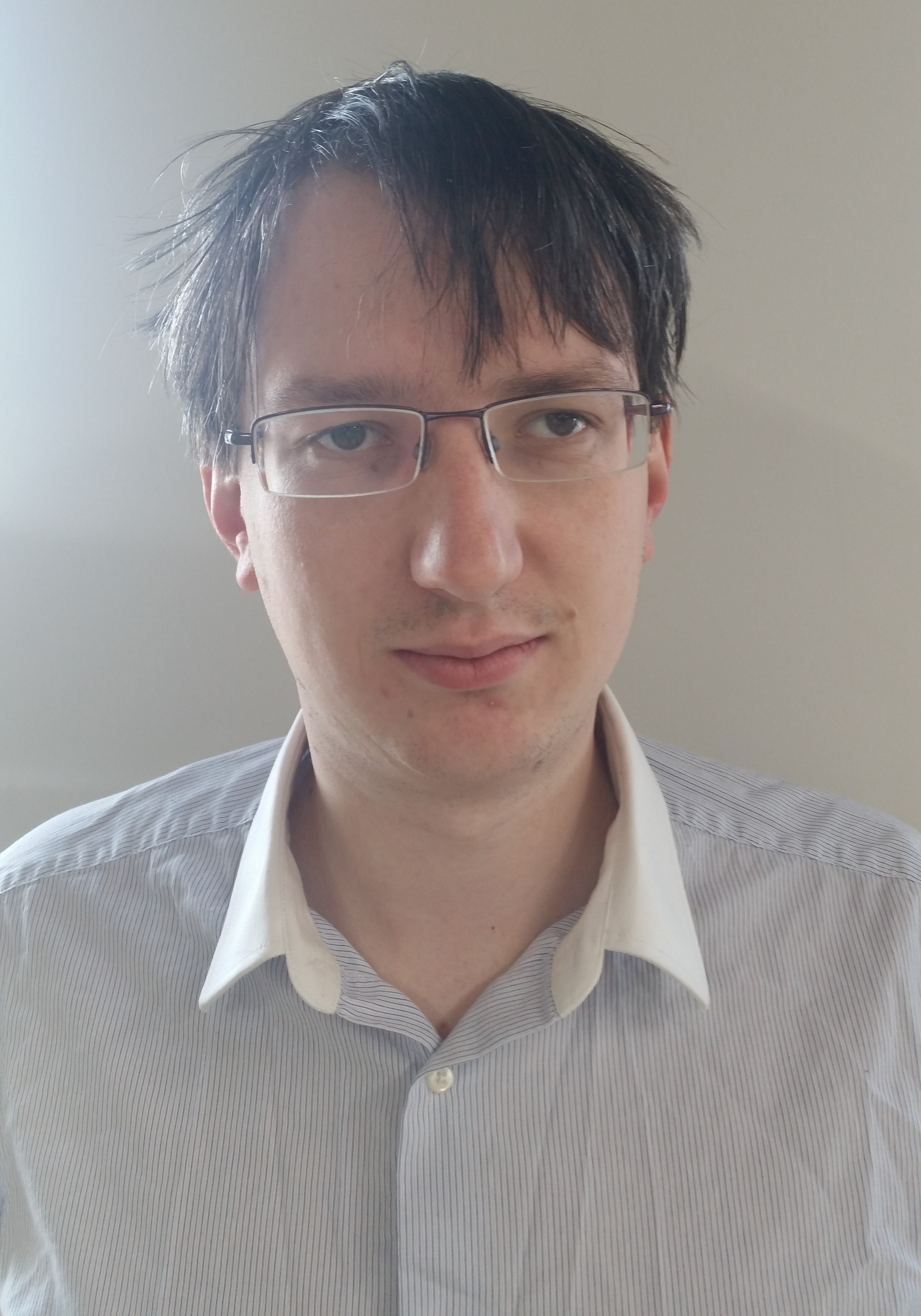}}]{Pawel Ladosz}
Pawel received the Meng (hons) degree in aerospace engineering in 2014 from Manchester University and a PhD degree in 2019 from Loughborough University, UK. He is currently Research Assistant Professor at Ulsan National Institute of Science and Technology (UNIST), South Korea. His research interests include reinforcement learning, robotics, optimization, deep learning and communication relays.
\end{IEEEbiography}

\begin{IEEEbiography}[{\includegraphics[width=1in,height=1.25in,clip,keepaspectratio]{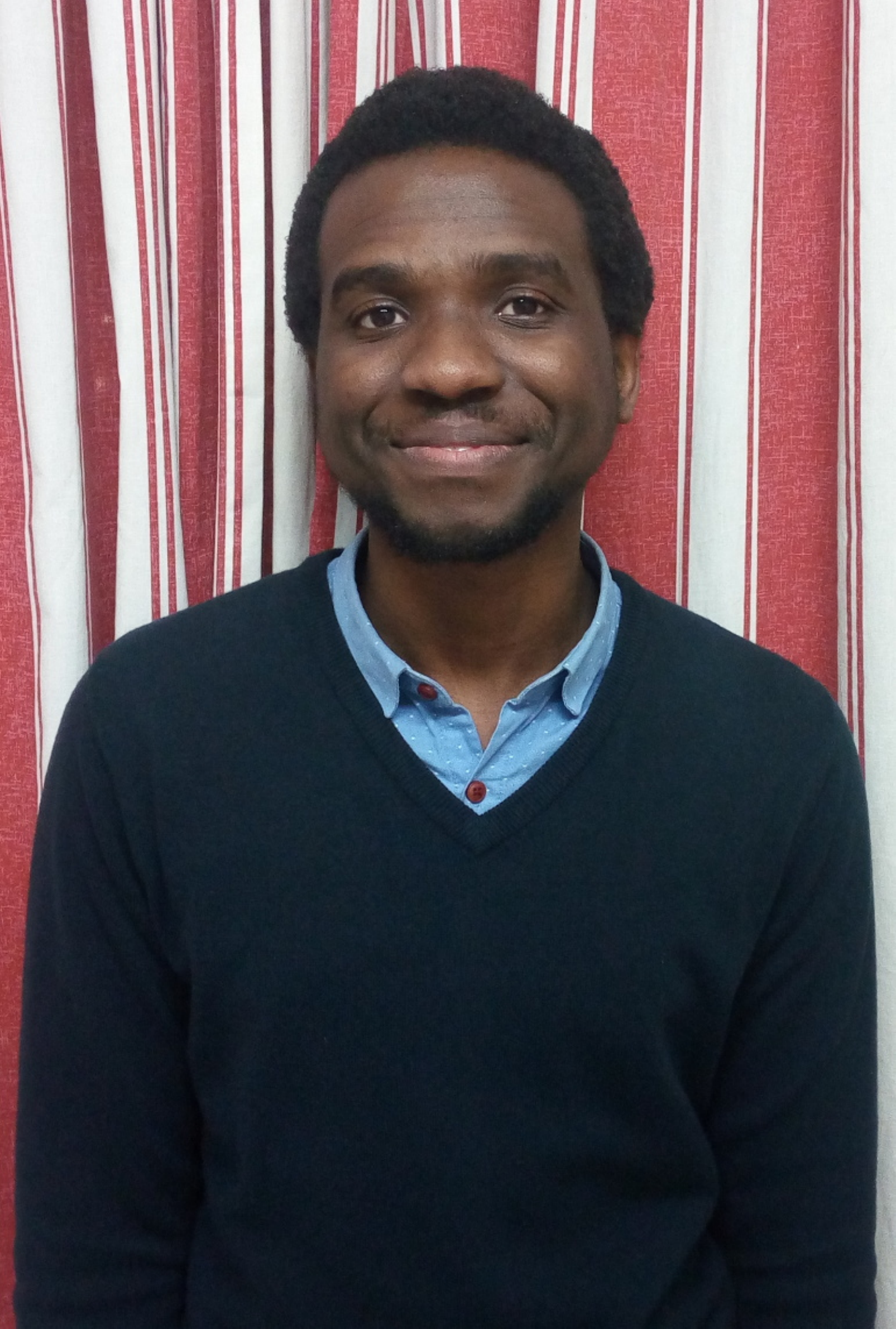}}]{Eseoghene Ben-Iwhiwhu}
Eseoghene received a MSc Computer Science degree from Coventry University in 2017. and he is currently a doctoral researcher at Loughborough University. His research is focused on the development of intelligent agents capable of lifelong learning and adaptation. To this end, his research investigates the development of neuromodulated neural networks, intersecting a number of research areas including; neuromodulation, neuroevolution, meta-reinforcement learning, deep reinforcement learning, and Hebbian learning.
\end{IEEEbiography}

\begin{IEEEbiography}[{\includegraphics[width=1in,height=1.25in,clip,keepaspectratio]{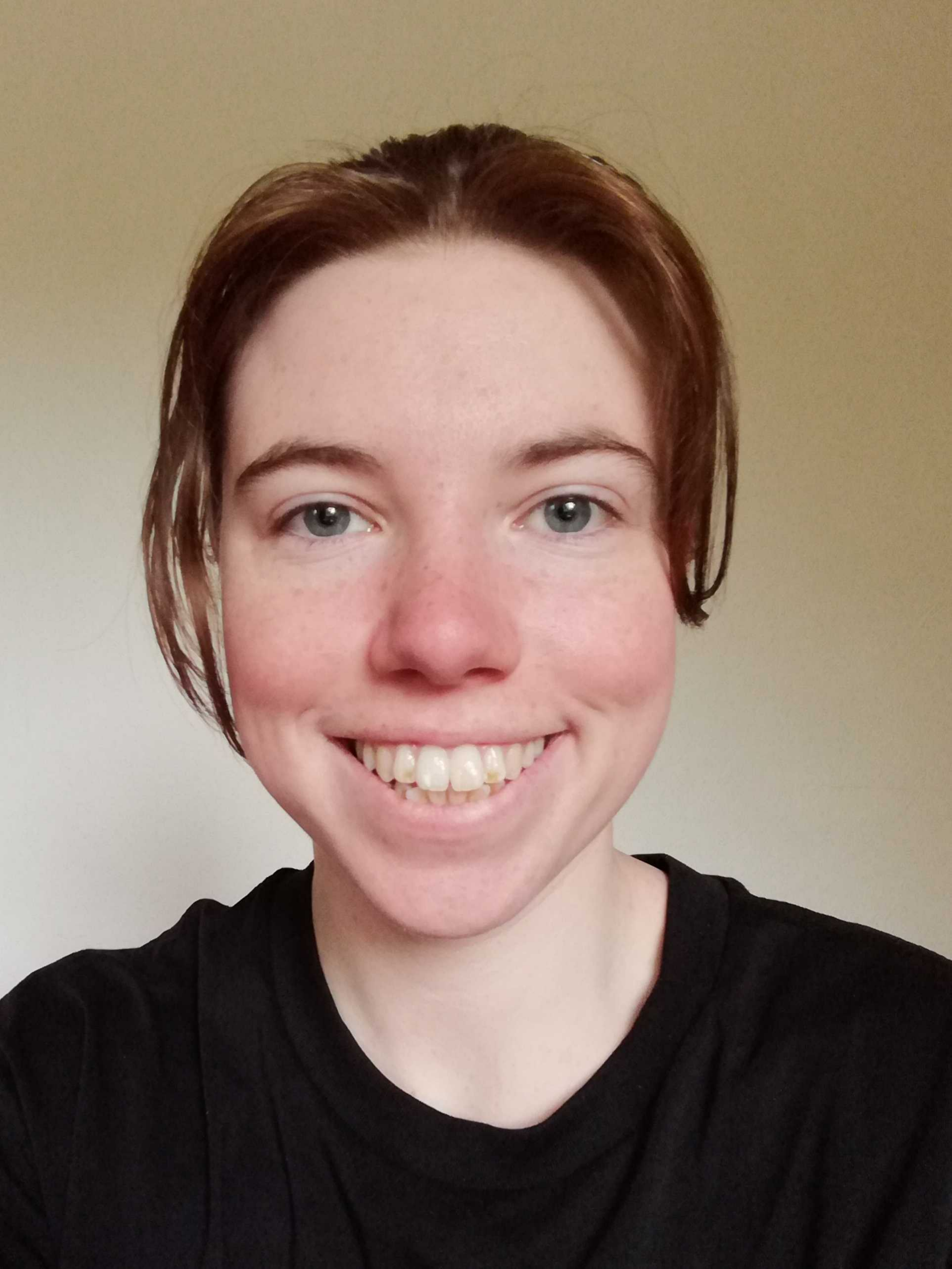}}]{Jeffery Dick}
Jeffery Dick is a graduate student based at Loughborough University. He earned his BsC in computer science and mathematics in 2019 from the same university, and uses his knowledge in both subjects in his current research.
\end{IEEEbiography}

\begin{IEEEbiography}[{\includegraphics[width=1in,height=1.25in,clip,keepaspectratio]{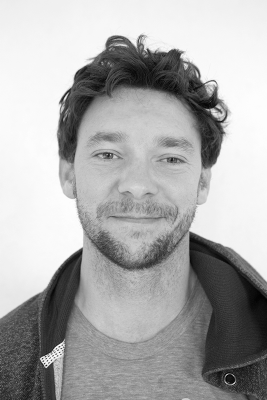}}]{Nicholas Ketz}
Nicholas Ketz received his B.A. in Physics from the University of Minnesota (2007) and Ph.D. in Computational Cognitive Neuroscience from University of Colorado Boulder (2016). He joined HRL Laboratories as a post-doc and eventually a research scientist within the Information and Systems Sciences Lab and the Human Machine Collaboration department. His work focuses on the intersection of human and artificial intelligence; studying biological systems and developing computational methods to advance the fields of machine learning, cognitive science and human neuroscience. Since 2018 he has been working on Lifelong Learning and domain adaptation within the context of autonomous driving.
\end{IEEEbiography}

\begin{IEEEbiography}[{\includegraphics[width=1in,height=1.25in,clip,keepaspectratio]{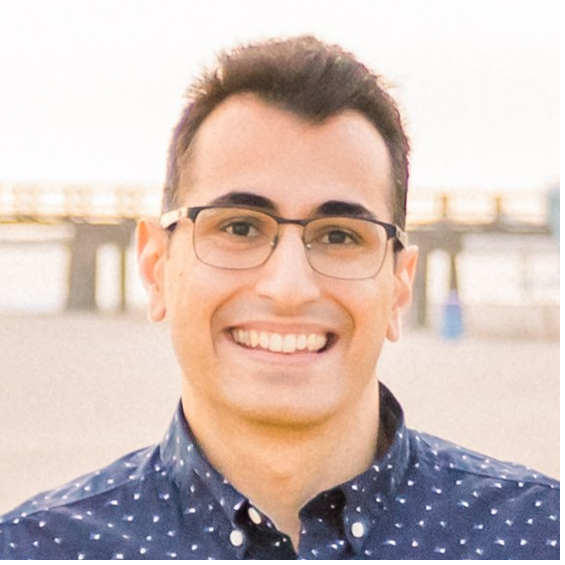}}]{Soheil Kolouri}
Soheil Kolouri is an Assistant Professor of Computer Science at Vanderbilt University, Nashville, TN, where he directs the Machine Intelligence and Neural Technologies (MINT) lab. He is broadly interested in machine learning, computer vision, and applied mathematics. He also has a standing interest in computational optimal transport and geometry. Before joining Vanderbilt, Soheil was a research scientist and a principal investigator at HRL Laboratories, Malibu, CA. He obtained his Ph.D. in Biomedical Engineering from Carnegie Mellon University, where he received the Bertucci Fellowship Award for outstanding graduate students from the College of Engineering in 2014 and the Outstanding Dissertation Award from the Biomedical Engineering Department in 2015.
\end{IEEEbiography}

\begin{IEEEbiography}[{\includegraphics[width=1in,height=1.25in,clip,keepaspectratio]{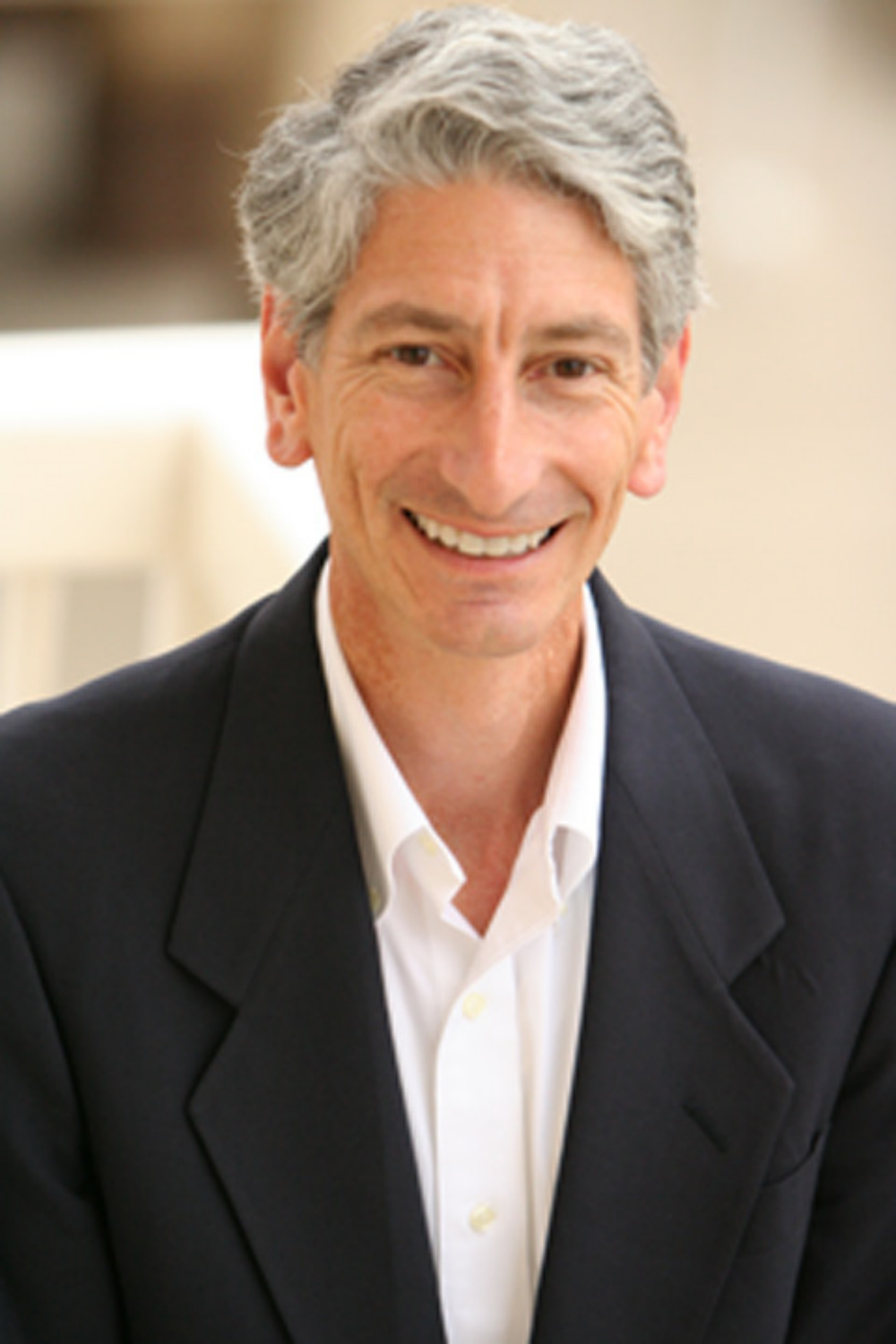}}]{Jeffrey L. Krichmar}
Jeffrey L. Krichmar received the B.S. degree in computer science from the University of Massachusetts at Amherst, Amherst, MA, USA, in 1983, the M.S. degree in computer science from The George Washington University, Washington, DC, USA, in 1991, and the Ph.D. degree in computational sciences and informatics from George Mason University, Fairfax, VA, USA, in 1997. Currently, he is a Professor with the Department of Cognitive Sciences and the Department of Computer Science, University of California at Irvine, Irvine, CA, USA. He has over 20 years’ experience in designing adaptive algorithms, creating neurobiologically plausible network simulations, and constructing brain-based robots whose behavior is guided by neurobiologically inspired models. His research interests include neurorobotics, embodied cognition, biologically plausible models of learning and memory, neuromorphic applications and tools, and the effect of neural architecture on neural function.
\end{IEEEbiography}

\begin{IEEEbiography}[{\includegraphics[width=1in,height=1.25in,clip,keepaspectratio]{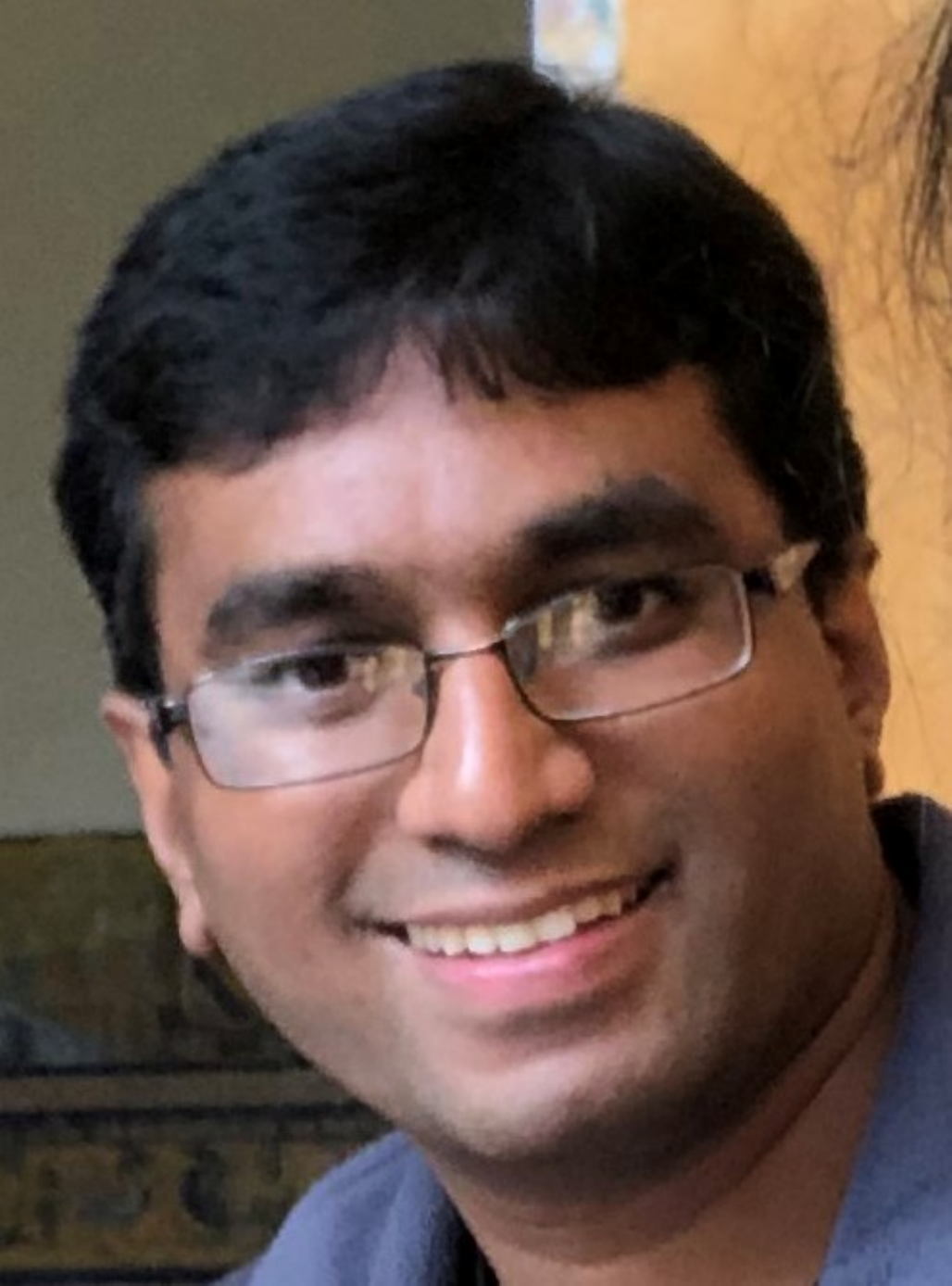}}]{Praveen Pilly}
Praveen K. Pilly received the B.Tech. degree in Electrical Engineering from the Indian Institute of Technology Madras, Chennai in 2003,  and the M.A. and Ph.D. degrees in Cognitive and Neural Systems from Boston University in 2008 and 2009, respectively. He was a Postdoctoral Research Associate and a Research Assistant Professor at Boston University from 2009 to 2012 and from 2012 to 2013, respectively. From 2013, he has been at HRL Laboratories where he is currently a Senior Research Scientist and the Leader of the Center for Human-Machine Collaboration. He was a Principal Investigator (PI) in the DARPA-funded RAM and RAM Replay programs. He was also the PI of an internal R\&D project, jointly funded by Boeing and General Motors, to develop real-world applications of HRL’s neuromorphic technology. He is currently the PI of the Super Turing Evolving Lifelong Learning ARchitecture (STELLAR) project in the DARPA-funded Lifelong Learning Machines (L2M) program to develop revolutionary brain-inspired continual learning and rapid adaptation algorithms for autonomous systems. His research interests are in learning and memory, cognitive and neural systems, artificial intelligence, and neurotechnology. He has 24 journal articles, 12 conference papers, and 25 US patents. Dr. Pilly is an Associate Editor for Frontiers in Human Neuroscience.
\end{IEEEbiography}

\begin{IEEEbiography}[{\includegraphics[width=1in,height=1.25in,clip,keepaspectratio]{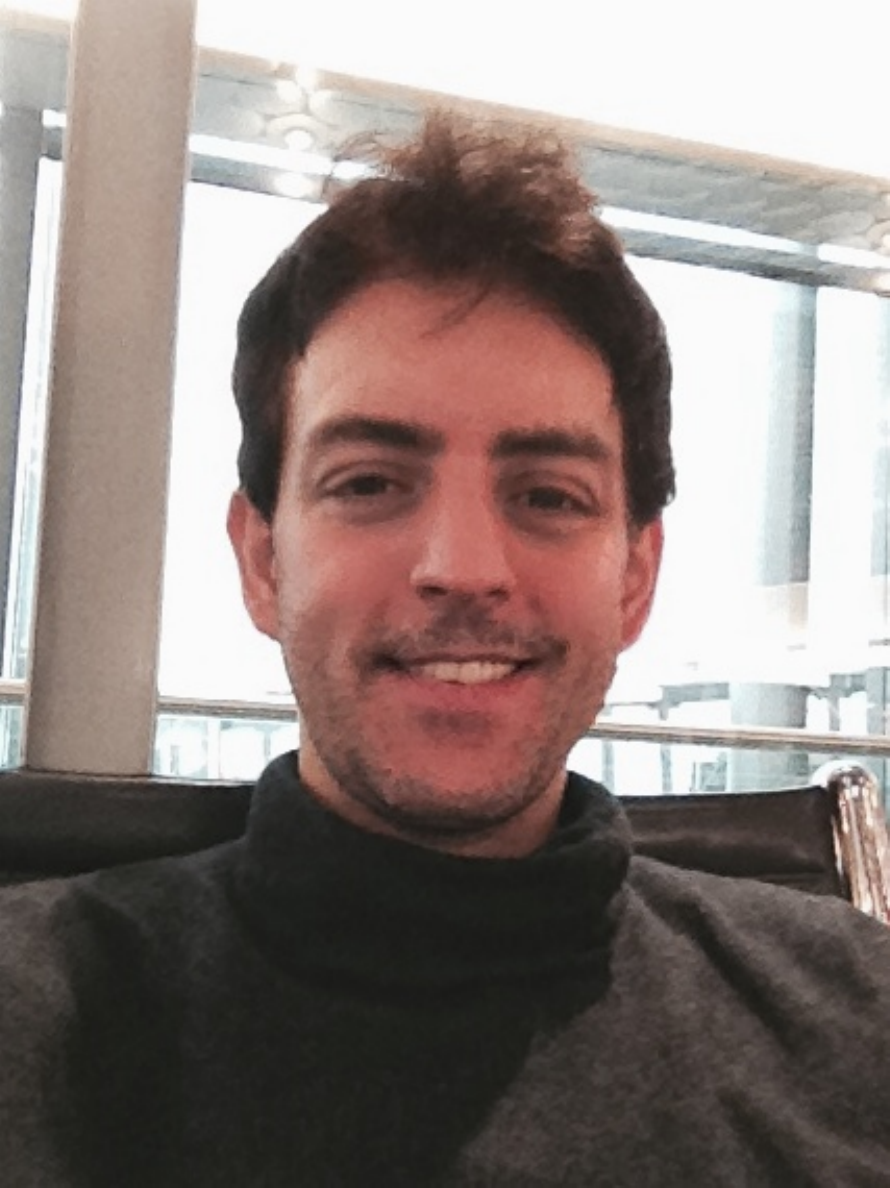}}]{Andrea Soltoggio}
Andrea Soltoggio received a B.Sc. and M.Sc. equivalent degrees in computer engineering from the Norwegian University of Science and Technology, Trondheim, Norway, and the Politecnico di Milano, Italy, in 2004, and a Ph.D. degree in computer science from the University of Birmingham, UK, in 2009.
Currently, he is a Senior Lecturer in Computer Science at Loughborough University, UK. His research interests include evolutionary computation, learning and plasticity in neural networks, lifelong learning, and broader aspects of cognition and intelligence.
\end{IEEEbiography}

\end{document}